\documentclass[nohyperref]{article}
\usepackage{hyperref}

\usepackage[accepted]{icml2022}

\usepackage[utf8]{inputenc}
\usepackage{array}
\usepackage{amsmath}
\usepackage{amssymb}
\usepackage{bm}
\usepackage{natbib}
\usepackage{graphicx}
\usepackage{booktabs}
\usepackage{setspace}
\usepackage{textcomp}
\usepackage{enumitem}
\usepackage{datetime2}
\usepackage{multirow}
\usepackage[font=small,labelfont=bf]{caption}
\DeclareMathOperator*{\argmin}{arg\,min}
\DeclareMathOperator*{\argmax}{arg\,max}
\usepackage{subcaption}
\usepackage{mathtools}
\usepackage{dirtytalk}

\usepackage[backgroundcolor=white]{todonotes}

\DeclareFontFamily{U}{cjheb}{}
\DeclareFontShape{U}{cjheb}{m}{n}{%
  <-11> s*[1.2] cjhblsm
  <11-> s*[1.2] cjhbltx
}{}
\newcommand{\cjhebfamily}{\fontencoding{U}\fontfamily{cjheb}\selectfont}
\DeclareTextFontCommand{\textcjheb}{\cjhebfamily}

\newcommand{\vv}{\mathbf{v}}
\newcommand{\vx}{\mathbf{x}}

\newcommand{\vbeta}{\boldsymbol{\beta}}

\newcommand{\vG}{\boldsymbol{G}}

\newcommand{\convergeD}{\xrightarrow[]{d}}

\newcommand{\vthe}{\boldsymbol{\theta}}
\newcommand{\mD}{\mathcal{D}}

\newcommand{\mJ}{\mathcal{J}}

\newcommand{\mA}{\mathcal{A}}

\newcommand{\I}{\mathbb{I}}

\newcommand{\name}[1]{\texttt{#1}}

\usepackage{soul}
\usepackage{xcolor}

\newcommand{\Mean}{{\mathbb{E}}}

\newcommand{\prob}{\mathbb{P}}

\newcommand{\vphi}{\boldsymbol{\phi}}
\newcommand{\vPhi}{\boldsymbol{\Phi}}

\newcommand{\vr}{\boldsymbol{r}}

\newcommand{\vone}{\boldsymbol{1}}
\newcommand{\vzero}{\boldsymbol{0}}

\newcommand{\vmu}{\boldsymbol{\mu}}

\newcommand{\mX}{\boldsymbol{X}}

\newcommand{\normal}{\mathcal{N}}

\usepackage{amsthm}
\newtheorem{thm}{Theorem}
\newtheorem{theorem}{Theorem}

\newtheorem{lemma}[thm]{Lemma}

\newtheorem{remark}{Remark}

\usepackage[ruled,vlined]{algorithm2e}

\makeatletter
\g@addto@macro\normalsize{%
\abovedisplayskip 7pt plus2pt minus5pt
\belowdisplayskip \abovedisplayskip 
\abovedisplayshortskip  0pt plus3pt%
\belowdisplayshortskip  4pt plus3pt minus3pt%
}
\makeatother

\newcommand{\var}{\mathrm{var}}

\begin{document}

\twocolumn[
\icmltitle{
Safe Exploration for Efficient Policy Evaluation and Comparison
}


\begin{icmlauthorlist}
\icmlauthor{Runzhe Wan}{to1}
\icmlauthor{Branislav Kveton}{to2}
\icmlauthor{Rui Song}{to1}
\end{icmlauthorlist}

\icmlaffiliation{to1}{Department of Statistics, North Carolina State University}
\icmlaffiliation{to2}{Amazon}

\icmlcorrespondingauthor{Rui Song}{rsong@ncsu.edu}

\icmlkeywords{Machine Learning, ICML}

\vskip 0.3in
]

\printAffiliationsAndNotice{} 

\begin{abstract}
High-quality data plays a central role in ensuring the accuracy of policy evaluation. 
This paper initiates the study of efficient and safe data collection for bandit policy evaluation. 
We formulate the problem and investigate its several representative variants. 
For each variant, we analyze its statistical properties, derive the corresponding exploration policy, and design an efficient algorithm for computing it. 
Both theoretical analysis and experiments support the usefulness of the proposed methods. 
\end{abstract}

\section{Introduction}
Bandit policies have been widely applied to areas including advertising \citep{bottou2013counterfactual}, search \citep{li2011unbiased} and healthcare \cite{zhou2017residual}. 
Before deploying a target policy, it is typically crucial to have an accurate evaluation  of its performance, which can subsequently provide valuable information for deployment decisions or model improvement.  
This is in general achieved by utilizing logged data, known as the \textit{off-policy evaluation (OPE)} problem. 

Although OPE has been studied extensively \cite{dudik2014doubly, li2015toward, swaminathan2016off, wang2017optimal, su2020doubly, kallus2021optimal, cai2021deep}, 
the existing works mainly focus on various estimators with a \textit{fixed} dataset. 
However, limited attention has been paid to the dataset itself, which plays a vital role in estimation accuracy. 
For instance, inverse probability weighting \citep[IPW, ][]{li2015toward}, one popular OPE method, has a key requirement that the logging policy has 
\say{full support}. 
This usually translates into requiring the logging policy to be close enough to the target policy \citep{sachdeva2020off, tran2021combining}. 
For direct method \citep[DM, ][]{dudik2014doubly}, it is also common that some feature directions are less explored in the logged dataset, which impacts the accuracy of this regression-based estimator. 


To improve the accuracy of policy evaluation, a natural idea is to actively collect high-quality data, and ideally, the data collection  rule should be tailored to the given  evaluation task. 
This problem is surprisingly unexplored. 
Moreover, due to the common safety concerns \cite{wu2016conservative, zhu2021safe}, 
it is of great practical importance to make sure the data collection rule is safe. 

\begin{figure}[!t]
\hspace{-.5cm}
\begin{subfigure}{.55\textwidth}
  \centering
  \includegraphics[width=\linewidth]{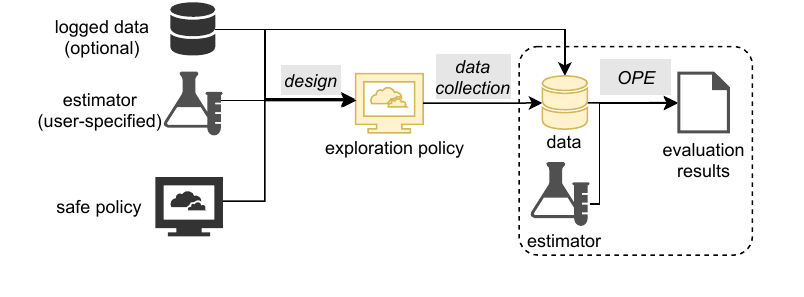}
\end{subfigure}%
\caption{SEPEC: designing a safe and efficient exploration policy to collect high-quality data for a given evaluation task. 
}
\label{fig:illustration}
\end{figure}

In this paper, we initiate the study of \textit{Safe Exploration for Policy Evaluation and Comparison (\name{SEPEC})}, the design of an efficient and safe exploration policy that collects a high-quality dataset for the given evaluation task. 
We  focus on non-adaptive policies, as they are simple to implement logistically. 
Indeed, in practice, the challenge of requiring investments (e.g., in infrastructures) for adaptive algorithms has been widely recognized \cite{zanette2021design, zhu2021safe}, especially before the value is proved via OPE. Our proposal is summarized in Figure \ref{fig:illustration}. 

Another possible approach to policy evaluation is being on-policy (i.e., following the target policy). 
However, the safety concern typically does not allow a direct deployment. 
In addition, perhaps surprisingly, the on-policy evaluation is not optimal under some setups (e.g., when the variances of different arms differ) or for some evaluation tasks (e.g., comparing the target policy with a baseline policy). 
See Section \ref{sec:MAB_IPW} for an example. 
Therefore, more careful analysis and design are required. 

As expected and shown later, our proposed solutions vary across different bandit setups, evaluation tasks, and value estimators. 
To shed light on this novel problem, we study its three representative variants, including multi-armed bandits (MAB) with IPW, contextual MAB (CMAB) with IPW, and linear bandit with DM. 
Our results can be extended to several other setups that are introduced as well. 
Finally, although other evaluation problems (e.g., estimating the value of the target policy in Section \ref{sec:PE}) can be addressed similarly to our work, for concreteness, we focus on \textit{policy comparison}, where we estimate the value difference between the target and baseline policies. 
This task is closely related to deployment decision making. 

\textbf{Contribution. } Our contributions can be summarized as follows. 
First, motivated by practical needs for improving policy evaluation accuracy and the common safety concerns, we propose and formulate the \name{SEPEC} framework. 
To the best of our knowledge, this is the \textit{first} work studying how to collect data for efficient policy evaluation, with or without  safety constraints. 
Second, we investigate its three representative variants thoroughly. 
We analyze their statistical properties, design tractable optimization problems, and provide efficient optimization algorithms. 
The situations with or without side information are both considered. 
Third, we theoretically prove the efficiency and optimality of our approach. 
Lastly, we demonstrate the superior performance of \name{SEPEC} through extensive experiments. 

\section{Objective and Existing Approaches}\label{sec:objective_and_baselines}
For concreteness, we first introduce the objective of \name{SEPEC} under MAB, and will extend it to other settings later. 
For any positive integer $M$, we denote the set $\{1, \dots, M\}$ by $[M]$. 
Let $\Delta_{K-1}$ be the $K$-dimensional simplex. 
In a $K$-armed bandit, 
we represent a policy by a vector $\pi = (\pi(1), \dots, \pi(K))^T \in \Delta_{K-1}$, where $\pi(a)$ is the probability that arm $a$ is pulled under policy $\pi$. 
After choosing an arm $A_t$, we receive its stochastic reward $R_t$. 
Let $r_a = \Mean[R_t | A_t = a]$ be the expected reward of arm $a$ and $\sigma^2_a = \var(R_t | A_t = a)$ be the variance of its rewards. 
Let $\vr = (r_1, \dots, r_K)^T$. 
The value of a policy is $V(\pi) = \sum_{a \in [K]} \pi(a) r_a$. 
Sometimes we write $V_{\vr}(\pi)$ to emphasize the dependency on $\vr$. 
We make the standard assumptions that $r_a \in [0,1]$ and $\sigma^2_a \le \sigma^2$ for some $\sigma > 0$. 
A problem instance is specified by $(\vr, \{\sigma_a\})$. 

We assume that we are given a \emph{target policy} $\pi_1$ and a safe \emph{baseline policy} $\pi_0$. 
We may also have side information, such as an existing dataset $\mD_0$. 
We focus on estimating the value difference $V(\pi_1) - V(\pi_0)$, although several other estimands (e.g., $V(\pi_1)$) can be addressed similarly. 
In practice, OPE methods are commonly used to estimate this lift and test its significance using existing data, which determines the deployment decision. 
However, the dataset is typically assumed as well explored, and 
little attention has been paid to how it arises. 
We aim to fill this gap. 
Specifically, given an \emph{exploration budget $T$}, a \emph{risk tolerance $\epsilon \in (0,1)$}, and a user-specified estimator that maps a dataset $\mD$ to a value estimate of policy $\pi$ as $\widehat{V}(\pi; \mD)$, 
we aim to design an \emph{exploration policy $\pi_e$} to collect a dataset $\mD_e$ of size $T$, while achieving the following two objectives simultaneously: 


\begin{itemize}
    \item \textbf{Safety}: Exploration should be safe, in the sense that $V_{\vr}(\pi_e) \ge (1-\epsilon) V_{\vr}(\pi_0)$ holds, either for all problem instances $\vr$ or with a high probability given side information. 
    \item  \textbf{Efficiency}: 
The exploration should be efficient, 
which we define as the minimization of $\var\big(\widehat{V}(\pi_1; \mD_0 \cup \mD_e) - \widehat{V}(\pi_0; \mD_0 \cup \mD_e)\big)$, i.e., 
the maximization of the evaluation accuracy with the given estimator. 
Specifically, we mainly focus on unbiased value  estimators, and in this case, the variance is closely related to many practical metrics, such as the mean squared error and the statistical power of testing $V(\pi_1) > V(\pi_0)$. 
See Section \ref{sec:testing} for details.
\end{itemize}

\textbf{Two existing approaches. }
To satisfy the safety constraint while exploring, the most popular practice \cite{thomas2015high, jiang2016doubly, slivkins2019introduction}, arguably, is to allocate an $\epsilon \in (0, 1)$ 
proportion of the budget to an exploration policy $b'\in \Delta_{K -1}$, 
and construct a \textit{mixture policy} $\pi_e = \epsilon b' + (1 - \epsilon) \pi_0$. 
The common choice of $b'$ is $\pi_1$ (i.e., on-policy) or the uniform distribution (i.e., random exploration). 
This approach, albeit being safe, could be inefficient. 
For example, $\pi_e$ is confined in a small policy class, and as we will show shortly, we can actually obtain a safe policy directly in a larger class via optimization. 
Besides, even within this policy class, $b'$ needs to be carefully designed and this problem alone is also underexplored. 


Another existing approach is \citet{zhu2021safe}, which also aims to collect high-quality bandit feedback in a safe manner. 
Our approach to handling the safety constraints is partially inspired by this paper. 
However, their targeted application is policy optimization, and hence they propose to maximize $\min_{a \in [K]} \pi_e(a)$, or in other words, they aim to collect data for evaluating \textit{all} policies jointly by reducing uncertainty uniformly. 
Therefore, their exploration policy is not tailored for evaluation, and will be less efficient without utilizing its specific structure. 
The \name{SEPEC} problem requires a more careful task-oriented analysis and design. 
A concrete example follows below. 



\textbf{Illustrative example.}
We start with analyzing the efficiency in a toy example without safety constraints. 
Consider $\pi_0 = (0.4, 0.1, 0.5)^T$ and $\pi_1 = (0.1, 0.4, 0.5)^T$. 
Then, one immediate observation is that, arm $3$ does not affect the estimate of the value difference $V(\pi_1) - V(\pi_0) = (0.1-0.4)r_1 + (0.4-0.1)r_2 + 0$, and so no budget should be spent on arm $3$ when exploring. 
Therefore, either following $\pi_1$, running A/B experiment (allocating $x\%$ budget to $\pi_1$ and $(100-x)\%$ to $\pi_0$), or using the uniform exploration of \citet{zhu2021safe} is clearly sub-optimal. 
The problem will become even more interesting when we maximize the efficiency with safety constraints, stochastic contexts, and generalization functions.


\section{Methodology}
In this section, we discuss our methodology by studying three OPE setups, including MAB with IPW,  CMAB with IPW, and linear bandits with DM. 
We choose these setups as they are representative, covering common features such as stochastic contexts and generalization functions. 
These discussions provide insights into this novel problem, and a few extensions will be introduced as well. 
In addition, we study both the case without side information, where we need to ensure safety under all instances, and the case with side information, which will enter the final estimator and  also can help us relax the safety constraint.

\begin{figure}[!h]
\centering
\includegraphics[width=\linewidth]{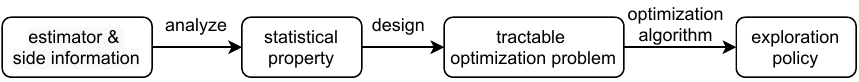}
\caption{Workflow for designing an exploration policy.}
\label{fig:workflow}
\end{figure}

For each setup, we start by analyzing the variance of the whole exploration and evaluation procedure. 
This raw objective is typically infeasible to optimize (e.g., it may involve unknown parameters). 
Therefore, we then design a tractable surrogate optimization problem, by solving which one can obtain an efficient policy. 
The optimization algorithm will be introduced at the end. 
We summarize the high-level idea in Figure \ref{fig:workflow}. 

\subsection{MAB with IPW}\label{sec:MAB_IPW}
We first apply \name{SEPEC} to evaluating an MAB policy using the IPW estimator, without side information. 
With a dataset $\mD_e = \{(A_t, R_t)\}_{t =1}^T$ collected following $\pi_e$, 
the IPW estimator is 
\begin{align*}
    \widehat{V}_{IPW}(\pi; \mD_e) = \frac{1}{T}\sum_{t=1}^T \frac{\pi(A_t)}{\pi_e(A_t)}R_t. 
\end{align*} 
Our objective can be written as
\begin{align*}
            &\argmin_{\pi_e \in \Delta_{K-1}} \var\big(\widehat{V}_{IPW}(\pi_1; \mD_{e}) - \widehat{V}_{IPW}(\pi_0; \mD_{e}) \big)\\
        &s.t. \;\; V_{\vr}(\pi_e) \ge (1-\epsilon) V_{\vr}(\pi_0), \forall \vr \in [0,1]^K. 
\end{align*} 
This problem is particularly challenging because (i) the dependency of the objective on $\pi_e$ is complex, 
and (ii) the safety constraint involves the unknown function $V$ and hence is not easy to guarantee. 

\textbf{Objective function.}
We aim to first transform the objective into a tractable form. 
Let $\pi_{\Delta} = \pi_1 - \pi_0$. 
We have 
\begin{align}
&T \times \var\Big(\widehat{V}_{IPW}(\pi_1; \mD_{e}) - \widehat{V}_{IPW}(\pi_0; \mD_{e}) \Big)\nonumber\\
&= T \times \var\Big(T^{-1} \sum_{t=1}^T \frac{\pi_{\Delta}(A_t)}{\pi_e(A_t)}R_t \Big)\nonumber\\
&= \var_{A_t \sim \pi_e}\Big(\frac{\pi_{\Delta}(A_t)}{\pi_e(A_t)}R_t \Big) \label{eqn:raw_MAB_obj}\\
&= 
\Mean_{A_t \sim \pi_e}\Big[\var(\frac{\pi_{\Delta}(A_t)}{\pi_e(A_t)}R_t | A_t)\Big]\nonumber\\
&\quad+ \var_{A_t \sim \pi_e}\Big[\Mean(\frac{\pi_{\Delta}(A_t)}{\pi_e(A_t)}R_t | A_t)\Big] \nonumber\\
&= 
\sum_{a \in [K]} \frac{\pi^2_{\Delta}(a)}{\pi_e(a)} \sigma^2_a 
+ \var_{A_t \sim \pi_e}\Big[\frac{\pi_{\Delta}(A_t)}{\pi_e(A_t)}r_{A_t}\Big]
 \label{eqn:0_MAB_objective_0}\\
 &= 
\sum_{a \in [K]} \frac{\pi^2_{\Delta}(a)}{\pi_e(a)} \sigma^2_a 
+ \sum_{a \in [K]} \frac{\pi^2_{\Delta}(a)}{\pi_e(a)} r^2_a
- c, 
 \label{eqn:0_MAB_objective_1}
\end{align}
where $c = (V(\pi_1) - V(\pi_0))^2$ is independent of $\pi_e$. 
The third equality is due to the law of total variance and the last one is due to $\var(X) = \Mean(X^2) - \Mean^2(X)$. 
We note that \eqref{eqn:0_MAB_objective_1} cannot be optimized directly, 
as $\sigma_a$'s and $r_a$'s are in general unknown. 
In fact, there does not exist a policy $\pi_e$ that achieves global optimum for all instances (see Appendix \ref{sec:Counter-example} for proof). 
However, this transformation provides insights into an efficient allocation rule. 
Without additional information, 
we consider an upper bound of the objective, by replacing $\sigma_a$'s and $r_a$'s with their joint upper bounds. 
Our relaxed objective is
\begin{equation}
    \begin{split}
    \argmin_{\pi_e  \in \Pi_{\Delta}} \sum_{a \in [K]} \frac{\pi^2_{\Delta}(a)}{\pi_e(a)},
    \end{split}
 \label{eqn:0_MAB_objective_no_constraint}
\end{equation} 
where the feasible set $\Pi_{\Delta} = \{\pi \in \Delta_{K - 1} : \pi(a) > 0 \; \text{if}\; \pi_{\Delta}(a) > 0 \}$ can be easily verified as convex. 
This ensures the \textit{positivity assumption} required by IPW \citep{li2015toward}. 
Although commonly assumed, this assumption has been found violated in many OPE applications \citep{sachdeva2020off, tran2021combining}, where IPW methods can fail catastrophically. 
This is particularly severe with contextual bandits or large action space. 
\name{SEPEC} proactively resolves this issue. 
Finally, without safety constraints, 
the solution of \eqref{eqn:0_MAB_objective_no_constraint} is $\pi_e(a) \propto |\pi_{\Delta}(a)|$. 
This result is intuitive: the larger the difference, the more we explore.





We make two remarks regarding the raw objective function \eqref{eqn:0_MAB_objective_0}.
First, perhaps surprisingly, by similar arguments, we can show that the on-policy strategy (i.e., $\pi_e = \pi_1$) is not optimal even in minimizing $\var\big(\widehat{V}(\pi_1; \mD_{e}))$ without safety constraints, when $\sigma_a$'s or $r_a$'s vary across arms. 
Instead, $\pi_e(a) \propto \pi_1(a) \times (r_a^2 + \sigma_a^2)$ is the most efficient one. 

Second, the two terms of \eqref{eqn:0_MAB_objective_0} correspond to the \textit{intrinsic uncertainty} from the reward noise after pulling  arms and the \textit{extrinsic randomness} from sampling arms. 
When $T \gg K$, 
it is possible to (approximately) remove the second term by considering a deterministic allocation. 
Specifically, given $\pi_e$, instead of sampling the $T$ data points, 
we can assign $N_a = T \times \pi_e(a)$ 
data points to the $a$th arm, which is known as the optimal continuous design in statistics \cite{lee1988constrained}.
However, in general, it is hard to guarantee that $N_a$'s are all integers, and one has to resort to randomized rounding to obtain an allocation $\{N_a\}$ that is close to  $\pi_e$. 
Although a few such techniques \cite{bouhtou2010submodularity, allen2017near} can yield small variance inflation when $T \gg K$, they are in general hard to generalize to broader setups such as contextual bandits, where the budget for each context is small and also stochastic. 
This is a setting of sharp difference with the standard experimental design literature. 
Therefore, we choose to sample from $\text{Multinomial}(T, \pi_e)$, which is a popular randomized rounding approach recently \cite{azizi2021fixed, zhu2021safe}. 

\textbf{Optimization without side information.}
With a tractable and appropriate surrogate objective, we are now ready to consider the safety constraint. 
The first observation is that $V_{\vr}(\pi_e) \ge (1-\epsilon) V_{\vr}(\pi_0), \forall \vr \in [0,1]^K$ is equivalent to $\pi_e(a) \ge (1-\epsilon) \pi_0(a), \forall a \in [K]$. 
This is because we need to ensure safety even in the worst case. 
Indeed, if there exists at least one arm $a$ such that $\pi_e(a) < (1-\epsilon) \pi_0$, we can always construct an  instance to violate the constraint, with $r_a = 1$ and $r_{a'} = 0, \forall a' \neq a$. Thus our optimization problem is
\begin{equation}
    \begin{split}
        &\argmin_{\pi_e  \in \Pi_{\Delta}} \sum_{a \in [K]} \frac{\pi^2_{\Delta}(a)}{\pi_e(a)} \\%
        &s.t. \;\; \pi_e(a) \ge (1-\epsilon) \pi_0(a), \forall a \in [K], 
    \end{split}
    \label{eqn:MAB_IPW_no_logged}
\end{equation}
and its solution is a safe policy. 
This problem is convex and hence can be efficiently solved by many  optimization algorithms \cite{boyd2004convex}. 



%

\textbf{Utilization of side information.}
Next, we study how to improve over  \eqref{eqn:MAB_IPW_no_logged} when there is side information, such as an existing dataset $\mD_0$ generated by a known behavior policy $\pi_b$. 
Typically $\pi_b$ is just $\pi_0$, but could be different. 
We assume  $\pi_b$ satisfies the positivity assumption; otherwise one can use IPW with a mixed propensity \cite{kallus2021optimal}. 
Specifically, 
we consider IPW with multiple logging policies \cite{kallus2021optimal}, 
and the original problem becomes
\begin{align*}
    &\argmin_{\pi_e \in \Pi_{\Delta}} 
    \var \Big(
    \sum_{t=1}^T \frac{\pi_{\Delta}(A_t)}{\pi_e(A_t)} R_t
    + {\sum_{(A_i, R_i) \in \mD_0}} \frac{\pi_{\Delta}(A_i)}{\pi_b(A_i)} R_i
    \Big)\\
    &s.t. \;\; V_{\vr}(\pi_e) \ge (1-\epsilon) V_{\vr}(\pi_0), \forall \vr \in \mathcal{C}_\delta(\mD_0). \nonumber
\end{align*}
Here $\mathcal{C}_\delta(\mD_0)$ is an $(1-\delta)$-confidence region on $\vr$ obtained from $\mD_0$, which can be constructed based on concentration inequalities. 
Alternatively, one can consider a Bayesian viewpoint when there is a prior on $\vr$ that summarizes domain knowledge, and $\mathcal{C}_\delta(\mD_0)$ is the credible region derived from the corresponding posteriors. We assume $\mathcal{C}_\delta(\mD_0)$ is a convex region, which is commonly the case following either approach.
See Appendix \ref{sec:appendix_constraint} for details.
The convexity allows finding the most violated constraint, for any fixed $\pi_e$, efficiently.
As such, the side information helps relax the safety constraint. 
Finally, we notice that for any $\vr$, $V_{\vr}(\pi_e) \ge (1-\epsilon) V_{\vr}(\pi_0)$ is equivalent to  $\vr^T(\pi_e - (1-\epsilon) \pi_0) \ge 0$, a linear constraint in $\pi_e$.

In addition, 
the objective      
is equal to
\begin{align*}
    \var \Big(
\sum_{t=1}^T \frac{\pi_{\Delta}(A_t)}{\pi_e(A_t)} R_t \Big)
+ \var\Big(\sum_{(A_i, R_i) \in \mD_0} \frac{\pi_{\Delta}(A_i)}{\pi_b(A_i)} R_i
\Big), 
\end{align*}
thanks to the independence between $\mD_e$ and $\mD_0$. 
The second term does not depend on $\pi_e$ and hence can be dropped in  optimization. 
In other words, perhaps surprisingly, the additional dataset $\mD_0$ does not \textit{directly} enter the objective. 
The main reason is that, every data point contributes independently to the IPW estimator. 
This may not hold for other estimators, as shown for DM in Section \ref{sec:Nonstochastic linear bandits}. 

\textbf{Optimization with side information. }
By similar arguments to \eqref{eqn:MAB_IPW_no_logged}, we propose an optimization problem
\begin{equation}
    \begin{split}
        &\argmin_{\pi_e \in \Pi_{\Delta}} \sum_{a \in [K]} \frac{\pi^2_{\Delta}(a)}{\pi_e(a)} \\%
        &s.t. \;\;  \vr^T(\pi_e - (1-\epsilon) \pi_0) \ge 0, \forall \vr \in \mathcal{C}_\delta(\mD_0). 
    \end{split}
    \label{eqn:MAB_IPW_w_logged}
\end{equation}
This problem is particularly challenging as the constraint implicitly contains an \textit{infinite} number of constraints. 
We note that, for any finite set $\Theta \subset \Delta_{k-1}$, 
the corresponding constraint $\vr^T(\pi_e - (1-\epsilon) \pi_0) \ge 0, \forall \vr \in \Theta$ is actually a finite number of linear constraints on $\pi_e$. 
Therefore, we propose to solve \eqref{eqn:MAB_IPW_w_logged} by adapting the cutting-plane method \cite{bertsimas1997introduction}. 
Specifically, the cutting-plane method iteratively adds more constraints to a finite set to approximate the convex feasible set, until no constraint in the original problem is violated. 
The convergence analysis is provided in \citet{boyd2007localization}. 
We present the algorithm details in Appendix \ref{sec:optimization_MAB}.





\begin{remark}\label{remark:alternative objective}
Recall that the raw objective \eqref{eqn:0_MAB_objective_1} involves the unknown parameters $\sigma_a$'s and $r_a$'s. 
We relax it using the upper bounds to guarantee the worst-case efficiency. 
When side information is available, 
alternatively, one can try to infer these parameters and consider a different objective function than \eqref{eqn:MAB_IPW_w_logged}, with a Bayesian or high-probability guarantee. 
We discuss this more in Appendix \ref{sec:obj_side_IPW}. 
\end{remark}

\subsection{CMAB with IPW}\label{sec:contextual_MAB_IPW}
In this section, we apply \name{SEPEC} to contextual MAB (CMAB). 
At every round $t$, we observe a context $\vx_t$ from a finite set of contexts $\mathcal{X}$, choose an arm $A_t \in [K]$, and then receive a stochastic reward $R_t$. 
We similarly define the mean reward ${r(a|\vx)= \Mean[R_t | A_t = a, \vx_t = \vx]}$, the mean reward vector $\vr(\cdot|\vx)\in [0, 1]^K$,  and the policy
$\pi(\cdot|\vx) \in \Delta_{K - 1}$. 
We also use $\vr = (\vr(\cdot|\vx))_{\vx\in\mathcal{X}}$ and $\pi = (\pi(\cdot|\vx))_{\vx\in\mathcal{X}}$ to denote the corresponding vector concatenated over $\mathcal{X}$. 
With $\pi_0$ and $\pi_1$ given, we define $\pi_{\Delta}$ as their difference. 
For design purpose, we make a common assumption \cite{wu2015algorithms, zhu2021safe} that the context distribution $p(\vx)$ is known, since there is typically a large dataset of contexts. 
Let $V_{\vr}(\pi|\vx) = \sum_{a \in [K]} \pi(a|\vx) r(a|\vx)$. 
The policy value is  $V_{\vr}(\pi) = \sum_{\vx\in \mathcal{X}}
p(\vx) V_{\vr}(\pi|\vx)$. 
Define $\Pi_{\Delta} = \{\pi : \pi(a|\vx) > 0 \; \text{if}\; \pi_{\Delta}(a|\vx) > 0, \forall a, \vx\}$. 
$\Pi_{\Delta}$ is a convex set by definition. 

For any dataset $\mD_e = \{(\vx_t, A_t, R_t)\}_{t=1}^T$ collected following a policy $\pi_e$, 
we consider the IPW estimator 
$\widehat{V}_{IPW}(\pi; \mD_e) = T^{-1}\sum_{t=1}^T \frac{\pi(A_t|\vx_t)}{\pi_e(A_t|\vx_t)}R_t$. 
Without side information, our problem is defined as 
\begin{equation}
    \begin{split}
        &\argmin_{\pi_e\in \Pi_{\Delta}} T \times \var\big(\widehat{V}_{IPW}(\pi_1; \mD_{e}) - \widehat{V}_{IPW}(\pi_0; \mD_{e}) \big)\\
        &s.t. \;\; V_{\vr}(\pi_e) \ge (1-\epsilon) V_{\vr}(\pi_0), \forall \vr \in [0,1]^{K |\mathcal{X}|}. 
    \end{split}
    \label{eqn:goal_CMAB}
\end{equation} 
We aim to construct an efficient and safe exploration policy as in the MAB (Section \ref{sec:MAB_IPW}). 
The main additional challenges come from the stochasticity in the contexts and that we need to solve the problem jointly over all contexts.



\textbf{No side information.}
We begin with studying the case without side information. 
We first note that our objective 
in \eqref{eqn:goal_CMAB} can be decomposed as 
\begin{align*}
&\Mean_{\vx_t \sim p(\vx)}
\big\{
\var[ \big(\frac{\pi_{\Delta} (A_t | \vx_t) }{\pi_e(A_t | \vx_t)}R_t \big) | \vx_t]
\big\}\\
&\quad+ 
\var_{\vx_t \sim p(\vx)}
\big\{ \Mean[ \big(\frac{\pi_{\Delta} (A_t | \vx_t) }{\pi_e(A_t | \vx_t)}R_t \big) | \vx_t]
\big\}. 
\end{align*}
The second term is equal to
\begin{align*}
    \var_{\vx_t \sim p(\vx)}\big[\sum_{a \in [K]} \pi_{\Delta} (a | \vx_t) r(a|\vx_t)\big], 
\end{align*}
does not depend on $\pi_e$, and hence can be dropped in  optimization. 
The first term can be regarded as the expectation of the MAB objective \eqref{eqn:raw_MAB_obj} over the context distribution. 

The second observation is that, 
the safety constraint is equivalent to $\pi_e(\cdot| \vx) \ge (1-\epsilon) \pi_0(\cdot| \vx), \forall \vx \in \mathcal{X}$. 
Otherwise, one can always construct a counter-example by setting $\vr(\cdot|\vx) = \vzero$, except for one context $\vx'$ where the constraint is violated. 
Then, we have $V(\pi_e) = p(\vx')V(\pi_e|\vx') < (1-\epsilon) p(\vx')V(\pi_0|\vx')
= (1-\epsilon) V(\pi_0)$, when $p(\vx') > 0$.

Therefore, by similar arguments as in Section \ref{sec:MAB_IPW} on deriving the surrogate objective, we propose the following tractable optimization problem
\begin{equation*}
    \begin{split}
        &\argmin_{\pi_e\in \Pi_{\Delta}} \sum_{\vx \in \mathcal{X}} \sum_{a \in [K]} p(\vx) \frac{\pi^2_{\Delta}(a| \vx)}{\pi_e(a| \vx)} \\
        &s.t. \;\; \pi_e(\cdot| \vx) \ge (1-\epsilon) \pi_0(\cdot| \vx), \forall \vx \in \mathcal{X},
    \end{split}
\end{equation*}
which can be solved by optimizing 
\begin{equation}
    \begin{split}
        &\argmin_{\pi_e(\cdot | \vx) \in \Pi_{\Delta}(\vx)} \sum_a \frac{\pi^2_{\Delta}(a| \vx)}{\pi_e(a| \vx)} \\
        &s.t. \;\; \pi_e(\cdot| \vx) \ge (1-\epsilon) \pi_0(\cdot| \vx), 
    \end{split}
    \label{eqn:CMAB_IPW_single}
\end{equation}
for each context $\vx \in \mathcal{X}$ separately. 
This is due to the additive form of IPW and the worst-case safety constraint. 
The optimization can be done similarly to \eqref{eqn:MAB_IPW_no_logged}. 

\textbf{Side information.}
The scenario considered above is arguably conservative. 
For instance, one may expect to solve all contexts jointly, which allows us to violate the constraint on some contexts and remedy on the others. 
We next consider the case with side information, e.g., an existing dataset $\mD_0 = \{(\vx_i, A_i, R_i)\}$, that allows us to do so. 

Regarding the objective function, 
by similar arguments to the MAB (Section \ref{sec:MAB_IPW}), 
$\mD_0$ only adds a $\pi_e$-independent term that can be dropped.  
However, $\mD_0$ can help us to relax the safety constraint. 
We can similarly construct a high-probability region $\mathcal{C}_\delta(\mD_0)$ such that $\vr \in \mathcal{C}_\delta(\mD_0)$ holds with probability at least $1 - \delta$. 
An efficient and safe policy can then be obtained from 
\begin{equation}
    \begin{split}
        &\argmin_{\pi_e\in \Pi_{\Delta}} 
        \sum_{\vx} p(\vx)
        \sum_a \frac{\pi^2_{\Delta}(a| \vx)}{\pi_e(a | \vx)} \\
        &s.t. \sum_{\vx \in \mathcal{X}}   \Big(\pi_e(\cdot | \vx) - (1 - \epsilon) \pi_0(\cdot | \vx) \Big) \\
        &\quad\quad\quad\quad 
       \times p(\vx) \vr(\cdot | \vx)^T \ge 0, \forall \vr \in \mathcal{C}_\delta(\mD_0), 
    \end{split}
\label{eqn:CMAB_IPW_2}
\end{equation} 
where we notice that, for every fixed $\vr$, the constraint is linear in $\pi_e$. 
Therefore, similar to \eqref{eqn:MAB_IPW_w_logged}, 
we propose to solve this problem by combining convex optimization algorithms with the cutting-plane method. 
See Appendix \ref{sec:optimization_MAB} for algorithm details. 





\subsection{Linear bandits with DM}\label{sec:Nonstochastic linear bandits}
Efficiency learning with a large action space typically relies on generalization functions, which also introduce interesting structures to the design of \name{SEPEC} algorithms. 
To shed light on this, we study linear bandits.  
At round $t$, we choose an arm $\vx_t$ 
from a set of arms $\mA$ with size $K$, and then receive a stochastic reward $R_t$. 
Each arm is represented by a $d$-dimensional vector.
Note that we overload notation and use $\vx_t$ to denote the feature vector of the pulled arm, instead of stochastic context. 
We assume that the expected reward of arm $\vx$ is $r(\vx) = \vx^T \vthe^*$ for some unknown parameter vector $\vthe^*$. 
For design purpose, 
we assume homoscedasticity, i.e., $\var(R_t | A_t) \equiv \sigma^2$. 
We use most notation introduced in the MAB. 
We remark that $V_{\vthe^*}(\pi) =  (\sum_{\vx \in \mA} \pi(\vx) \vx)^T \vthe^* = \bar{\vphi}^T_{\pi} \vthe^*$, where $\bar{\vphi}_{\pi} = \sum_{\vx \in \mA} \pi(\vx) \vx$ is the mean feature direction of $\pi$. We assume that $V_{\vthe^*}(\pi_0) \ge 0$. 

To estimate the value of a policy $\pi$, 
we consider the direct method \citep[DM, ][]{dudik2014doubly}. 
With a dataset $\mD = \{(\vx_i, R_i)\}$, 
DM first estimates $\vthe^*$ via least-square regression as $\hat{\vthe} = (\sum_{(\vx_i, R_i) \in \mD} \vx_i \vx_i^T)^+ (\sum_{(\vx_i, R_i) \in \mD} \vx_i R_i)$, and then 
plugs-in $\hat{\vthe}$ in the value definition to construct $\widehat{V}_{DM}(\pi; \mD)  = \sum_{\vx \in \mA} \pi(\vx) (\vx^T \hat{\vthe})= \bar{\vphi}^T_{\pi} \hat{\vthe}$. 


\textbf{Objective function.}
We pull $T$ arms $\vPhi = (\vx_1, \dots, \vx_T)^T$ to collect a  dataset $\mD_{\vPhi}$, which is stochastic and depends on $\vPhi$. 
In this section, we directly consider the case with side information, since it does not make a  significant difference in derivations. 
We assume we have an existing dataset $\mD_0$ and denote the corresponding feature matrix as $\vPhi_0$, which we assume is full-rank.  
Its existence simplifies our exposition. 
Alternatively, one can always use forced exploration to form a basis, or consider regularized least squares. 
Towards the goal of variance minimization, we note that
\scalebox{0.92}{\parbox{\linewidth}{%
\begin{align}
&\var_{\vPhi \sim \pi_e} \Big(\widehat{V}_{DM}(\pi_1; \mD_0 \cup \mD_{\vPhi}) - \widehat{V}_{DM}(\pi_0;\mD_0 \cup \mD_{\vPhi}) \Big)\nonumber\\
&= 
\Mean_{\vPhi \sim \pi_e}\Big[\var \Big(\widehat{V}_{DM}(\pi_1; \mD_0 \cup \mD_{\vPhi}) - \widehat{V}_{DM}(\pi_0; \mD_0 \cup \mD_{\vPhi}) | \vPhi \Big)
 \Big]\nonumber\\
&\;+ \var_{\vPhi \sim \pi_e}\Big[ \Mean \Big(\widehat{V}_{DM}(\pi_1; \mD_0 \cup\mD_{\vPhi}) - \widehat{V}_{DM}(\pi_0; \mD_0 \cup\mD_{\vPhi}) | \vPhi \Big)
\Big]\nonumber\\
&= 
\sigma^2
\Mean_{\vPhi \sim \pi_e}\Big[\bar{\vphi}^T_{\Delta \pi}
     (\vPhi^T\vPhi+ \vPhi_0^{T}\vPhi_0)^{-1}
    \bar{\vphi}_{\Delta \pi}
 \Big]\nonumber\\
&\quad\quad + \var_{\vPhi \sim \pi_e}[V(\pi_1) {-} V(\pi_0)]\nonumber\\
&= 
\sigma^2
\Mean_{\vPhi \sim \pi_e}\Big[\bar{\vphi}^T_{\Delta \pi}
     (\vPhi^T\vPhi+ \vPhi_0^{T}\vPhi_0)^{-1}
    \bar{\vphi}_{\Delta \pi}
 \Big] + 0. 
    \label{eqn:raw_LB}
\end{align}
}} 
The generalization function introduces interesting structures. 
For example, unlike in IPW, $\mD_0$ also enters our optimization objective, since actions are related through their features. 
Intuitively, we should spend less budget on those extensively explored directions. 
Moreover, unlike IPW, the second term of \eqref{eqn:raw_LB} is zero. 
This is because, conditioned on any set of sampled arms $\vPhi$, DM estimator is always unbiased and hence its conditional expectation is independent with $\vPhi$. 




The optimization of \eqref{eqn:raw_LB} is very challenging, since it involves an expectation over the inverse of a random matrix. 
We notice that this objective is related to the G-optimal experiment design problem \cite{shah2012theory}, where we aim to minimize the maximum uncertainty 
by $\argmin_{\vPhi} \max_{\Tilde{\vx} \in \mathcal{X}} \Tilde{\vx}^T (\vPhi^T\vPhi)^{-1} \Tilde{\vx}$. 
For this problem, since a direct optimization is still NP-hard, 
it is common to solve its continuous relaxation \cite{shah2012theory, zhu2021safe} 
$\argmin_{\pi_e} \max_{\Tilde{\vx} \in \mathcal{X}} \Tilde{\vx}^T  (T\sum_{\vx \in \mathcal{A}} \pi_e(\vx) \vx \vx^T)^{-1} \Tilde{\vx}$, and then apply certain randomized rounding methods to construct $\{\vx_1, \dots, \vx_T\}$ from the distribution $\pi_e$.

Motivated by the good property of this approximation \cite{shah2012theory, lattimore2020bandit}, 
we consider a similar relaxation for our problem as 
\begin{align*}
\argmin_{\pi_e \in \Pi'_{\Delta}}  \bar{\vphi}^T_{\Delta \pi}
     (T\sum_{\vx \in \mathcal{A}} \pi_e(\vx) \vx \vx^T + \vPhi_0^{T}\vPhi_0)^{-1}
    \bar{\vphi}_{\Delta \pi},  
\end{align*} 
where $\Pi'_{\Delta} = \{\pi \in \Delta_{K - 1} {:} \sum_{\vx \in \mathcal{A}} \pi(\vx) \vx \vx^T \, \text{is invertible} \}$ is a convex set by definition. 
In Section \ref{sec:analysis}, we prove that this objective is actually the \textit{asymptotic variance} of the DM estimator.

\textbf{Optimization.}
After obtaining a tractable objective function, we next study solving the exploration policy with the safety constraint. 
Again, we can construct a high-probability region $\mathcal{C}_\delta(\mD_0)$ for $\vthe$ in either a frequentist way or a Bayesian way. 
With either approach, the region is typically an ellipsoid and hence convex, which we assume hereinafter.
See Appendix \ref{sec:appendix_constraint} for some examples.
The convexity allows finding the most violated constraint, for any fixed $\pi_e$, efficiently.
The exploration policy can be obtained from
\begin{equation}
    \begin{split}
&\argmin_{\pi_e \in \Pi'_{\Delta}} \;
\bar{\vphi}^T_{\Delta \pi}
    (T\sum_{\vx \in \mathcal{A}} \pi_e(\vx) \vx \vx^T+ \vPhi_0^{T}\vPhi_0)^{-1}
    \bar{\vphi}_{\Delta \pi}\\
&s.t. \; V_{\vthe}(\pi_e) \ge (1-\epsilon)V_{\vthe}(\pi_0)  , \forall \vthe \in \mathcal{C}_\delta(\mD_0). 
    \end{split}
\label{eqn:nonstochastic linear bandit with side info}
\end{equation} 
The constraint is equivalent to $ \big(\pi_e - (1-\epsilon) \pi_0\big)^T \mX \vthe \ge  0, \forall \vthe \in \mathcal{C}_\delta(\mD_0)$, 
where $\mX$ is the feature matrix obtained by stacking vectors in $\mA$.
Therefore, Problem \eqref{eqn:nonstochastic linear bandit with side info} is convex. 
For a finite set of constraints, we adapt the popular Frank–Wolfe (FW) algorithm \citep{frank1956algorithm}, which is a projection-free algorithm with good convergence guarantees \citep{jaggi2013revisiting}. 
To handle the infinite number of constraints, we combine the FW algorithm with the cutting-plane method. 
See Appendix \ref{sec:optimization_LB} for details.


\subsection{Policy evaluation}\label{sec:PE}

For concreteness, we choose the policy comparison problem to present our methodology. 
We would like to emphasize that all discussions are equally applicable to evaluating the value of a single policy, i.e., estimating $V(\pi_1)$. 
To see this, note that due to linearity, all discussions on the objective functions still hold, e.g.,  by replacing $\pi_{\Delta}$ in \eqref{eqn:raw_MAB_obj} with $\pi_1$ or $\Bar{\phi}_{\Delta \pi}$ in \eqref{eqn:raw_LB} with $\Bar{\phi}_{\pi_1}$. 
Moreover, the safety constraint is independent of the estimand. 
Therefore, the optimization problem can be formulated and solved in almost the same manner as for policy comparison. 


\subsection{Extensions}\label{sec:extensions}
As we expect and also observed, the specific solution for \name{SEPEC}  vary across different bandit setups, evaluation tasks and value estimators. 
To initiate the study of this novel area, 
we considered three representative variants covering MAB, contextual problems, generalization functions, IPW and DM. 
Our discussion, derivations and algorithms can be extended to at least the following problems. 

First, for MAB, the DM estimator is equivalent to an alternative IPW-form estimator, which we analyze in Appendix \ref{sec:MAB_DM}. 
Second, for stochastic contextual linear bandits 
where the stochastic context $\vx_t$ and action $A_t$ together generate a feature vector $\vphi_{\vx_t, A_t}$ that determines the reward linearly, 
our analysis for IPW and linear bandits can be combined and extended. 
See Appendix \ref{sec:contextual_LB} for details. 
Third, 
linear bandits with IPW are usually referred to as the pseudo-inverse estimator \cite{swaminathan2016off}, and its contextual version is particularly useful on some structured problems, such as slate recommendation. 
Our analysis in Section \ref{sec:Nonstochastic linear bandits} can be extended to these problems. 
See Appendix \ref{sec:PI} for details. 
Lastly, another popular value estimator is the doubly robust (DR) estimator  \cite{dudik2014doubly}, which combines IPW and DM. 
Discussions in Section \ref{sec:contextual_MAB_IPW} can be similarly applied to minimize the asymptotic variance when DR is used as the value estimator. 
See Appendix \ref{sec:DR} for details. 

\section{Theoretical Analysis}\label{sec:analysis}
In this section, we provide theoretical analysis for \name{SEPEC}. 
Since our optimization problems can all be solved \textit{exactly}, the safety constraint can be satisfied as promised. 
Therefore, we will focus on the efficiency maximization problem in Section \ref{sec:Efficiency}. 
All proofs are deferred to Appendix \ref{sec:proofs}. 
The connection between our objective and the testing power is discussed in Section 
\ref{sec:testing}. 

\subsection{Efficiency}\label{sec:Efficiency}
We first study MAB with IPW. 
The same conclusions can be established for CMAB under similar conditions. 
To provide insights, we first consider the case without side information and safety constraints to derive an explicit form of the overall variance. 
Denote $\var(\widehat{V}_{IPW}(\pi_1;\mD_e) - \widehat{V}_{IPW}(\pi_0; \mD_e))$ as $\nu(\pi_e; \vr, \{\sigma_a\})$ when the true parameters are $\vr$ and $\{\sigma_a\}$. 
Denote the solution of \eqref{eqn:0_MAB_objective_no_constraint} as $\pi_e^*$. 
\begin{lemma}\label{lemma:ipw_explicit_form}
The overall variance $\nu(\pi_e^*; \vr, \{\sigma_a\})$ is 
\scalebox{0.9}{\parbox{\linewidth}{%
\begin{align*}
\frac{1}{T} \big[
\sum_{a \in [K]} |\pi_{\Delta}(a)| {\times}
\sum_{a \in [K]} |\pi_{\Delta}(a)| (\sigma^2_a + r^2_a)
- \big(V(\pi_1) - V(\pi_0)\big)^2
\big]. 
\end{align*}
}}
\end{lemma} 
As expected, the variance decays at rate $T^{-1}$. 
The result is intuitive as it depends on how different the two policies are (i.e., $\sum_a |\pi_{\Delta}(a)|$) and if the difference is more significant on those more important arms (with larger $\sigma^2_a$ or $r^2_a$). 

Next we study the efficiency.
As mentioned in Section \ref{sec:MAB_IPW}, the objective involves unknown parameters and 
there is no policy that always dominates (see Appendix \ref{sec:Counter-example} for proof). 
Therefore, to provide insights, we study the minimax performance and the average performance. 
For the former, we consider the set of instance $\mathcal{I} = \{(\vr, \{\sigma_a\}): \vzero \le \vr \le \vone, 0 \le \sigma_a \le \sigma, \forall a \}$. 
For the latter, we consider any instance distribution $Q$ such that 
$\Mean_Q(r_a)$, $\var_Q(r_a)$ and $\Mean_Q(\sigma^2_a)$ all have fixed values across the $K$ arms. 
For either Problem  \eqref{eqn:0_MAB_objective_no_constraint}, \eqref{eqn:MAB_IPW_no_logged} or \eqref{eqn:MAB_IPW_w_logged}, 
denote the solution (i.e., our policy) as $\pi_e^*$ and the set of feasible policies as $\Pi$. 
Note that $\Pi$ can be different with $\Pi_{\Delta}$, since $\Pi$ may be under safety constraints. 
\name{SEPEC} enjoys the following nice properties. 
\begin{theorem}\label{thm:mab}
The policy $\pi_e^*$ is minimax optimal, i.e., 
\begin{align*}
    \pi_e^* \in \argmin_{\pi_e \in \Pi} \max_{(\vr, \{\sigma_a\})  \in \mathcal{I}} \; \nu(\pi_e; \vr, \{\sigma_a\}). 
\end{align*}
The policy $\pi_e^*$ is also the most efficient on average, i.e., 
\begin{align*}
\pi_e^* \in \argmin_{\pi_e \in \Pi} \Mean_{(\vr, {\{\sigma_a\}}) \sim Q} \; \nu(\pi_e; \vr, \{\sigma_a\}). 
\end{align*}
\end{theorem}
Finally, we study linear bandits with DM. 
For any policy $\pi_e$, with logged data $\mD_0$, we denote the ultimate objective \eqref{eqn:raw_LB} as $\nu^*(\pi_e; \mD_0)$ 
and the surrogate objective 
in \eqref{eqn:nonstochastic linear bandit with side info}
as $\nu(\pi_e;\mD_0)$. 
We have the following promised result. 
\begin{theorem}\label{thm:LB}
For any policy $\pi_e \in \Pi'_{\Delta}$, 
we have $T \times |\nu(\pi_e;\mD_0) - \nu^*(\pi_e; \mD_0)| \to 0$ as $T$ grows. 
In other words, the surrogate objective $\nu(\pi_e;\mD_0)$ is the asymptotic variance of 
$\widehat{V}_{DM}(\pi_1;\mD_0 \cup \mD_e) - \widehat{V}_{DM}(\pi_0; \mD_0 \cup \mD_e)$. 
\end{theorem}
In this sense, we regard our surrogate problem as a good proxy, which we solve \textit{exactly}. 
Moreover, notice that $\nu$ and $\nu^*$ are both continuous. 
Therefore, suppose the optimum is in a compact set where the convergence is uniform, then 
we know $T \times |\nu^*(\pi_e^*;\mD_0) - \min_{\pi_e} \nu^*(\pi_e; \mD_0)| \to 0$,  i.e., $\pi_e^*$ is \textit{asymptotically optimal}. 
We leave the finite-sample analysis for future research and investigate it experimentally here. 





\subsection{Connection with hypothesis testing}\label{sec:testing}
In real applications, one important task is to test whether or not the target policy $\pi_1$ is significantly better than $\pi_0$, which impacts deployment of $\pi_1$. 
Thus we investigate the relationship between the power of such a test and our objective $\var\big(\widehat{V}(\pi_1; \mD_{e}) - \widehat{V}(\pi_0; \mD_{e}) \big)$. 
For simplicity, we discuss MAB with IPW and without side information. 
A similar connection can be established for other setups considered in this paper (see Appendix \ref{sec:other_testing}). 
Specifically, we consider the test
\begin{align*}
    H_0: V(\pi_1) \le V(\pi_0) \quad \text{v.s.} \quad H_1: V(\pi_1) > V(\pi_0). 
\end{align*} 
Denote our variance objective \eqref{eqn:raw_MAB_obj} by $\sigma(\pi_e)$  and let $\hat{\sigma}(\pi_e)$ be a consistent estimator with $\mD_e$ (typically the plug-in estimator with sample variance). 
By the central-limit theorem and the Slutsky’s lemma, we have 
\begin{align*}
    \hat{\sigma}(\pi_e)^{-1} \sqrt{T}
    &\Big[
    \big(\widehat{V}(\pi_1; \mD_{e}) 
    - \widehat{V}(\pi_0; \mD_{e}) \big)\\[-1ex]
    &\quad- \big(V(\pi_1) - V(\pi_0)\big)\Big]
    \overset{d}{\to} \normal(0, 1), 
\end{align*}
when $T$ grows.  
Therefore, a popular (asymptotically) $\alpha$-level test \cite{cai2020validation} is to reject the null when 
\begin{align*}
\sqrt{T}\hat{\sigma}(\pi_e)^{-1}
\big(\widehat{V}(\pi_1; \mD_{e}) 
- \widehat{V}(\pi_0; \mD_{e}) \big) \ge z_{1-\alpha}, 
\end{align*}
where $z_{1-\alpha}$ is  the upper $\alpha$th quantile of the standard normal distribution. 
For any $\Delta > 0$, the (asymptotic) power under the local alternative $H_1 : V(\pi_1) - V(\pi_0) = \Delta / \sqrt{T}$ 
is hence 
$1 - \Phi\big(z_{1-\alpha} - \sigma(\pi_e)^{-1} \Delta\big)$, where $\Phi(\cdot)$ is the cumulative distribution function of standard Gaussian. 
Since the other terms are all constant, it is clear that $\sigma(\pi_e)$ determines the power. 
Therefore, with a fixed budget, one can improve the power by designing a better policy $\pi_e$ to reduce $\sigma(\pi_e)$, which corresponds to our objective. 







\section{Experiments}\label{sec:expt}
In this section, we compare the empirical performance of various methods. 
We focus on three metrics: 
(i) the root mean square error (RMSE) of estimation, which quantifies the  efficiency; 
(ii) the power of detecting $V(\pi_1) > V(\pi_0)$ when it holds, which quantifies the downstream impact; 
and (iii) the loss from safety violation $\max(0, T(1-\epsilon){\times}V(\pi_0) - \sum_{t=1}^T \Mean(R_t | A_t) )$, which quantifies the risk. 

We compare our method \name{SEPEC} with several baselines. 
The first two are introduced in Section \ref{sec:objective_and_baselines}. 
To recap, the mixture policy $\pi_e = \epsilon \pi_1 + (1-\epsilon) \pi_0$ is a common way to run safe policy comparison, while the Safe Optimal Design (SafeOD) proposed in \citet{zhu2021safe} ignores task-specific structure. 
Besides, we study the performance of \textit{uniform sampling}, which is a common exploration policy; and \textit{A/B test}, which allocates a half of the budget to $\pi_1$ and the rest to $\pi_0$, and is a common practice for policy comparison.  
Finally, we consider the variant of \name{SEPEC} that does not consider the safety constraint during optimization. 




\newcommand{\sharedwidth}{0.5}
\newcommand{\sharedheight}{3.2cm}
\newcommand{\hbetween}{0.1cm}
\newcommand{\vbetween}{-0.5cm}
\newcommand{\hleft}{-0.3cm}

\begin{figure}[!t]
\hspace{\hleft}
\begin{subfigure}{\sharedwidth\textwidth}
  \centering
  \includegraphics[width=\linewidth, height = \sharedheight]{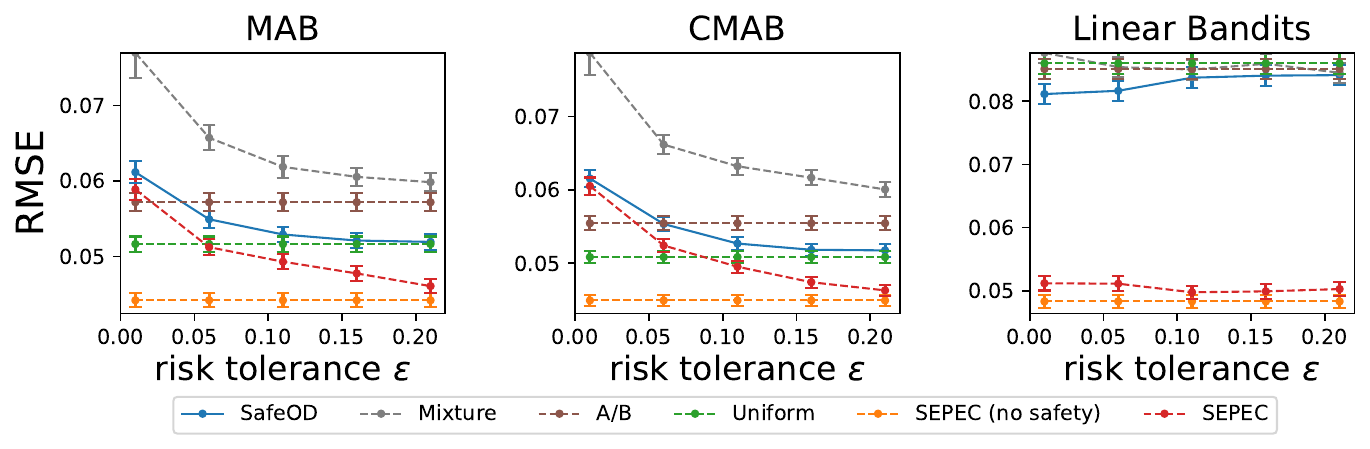}
    \vspace{\vbetween}
  \caption{
  Estimation efficiency measured by RMSE. 
  }
\end{subfigure}%

\hspace{\hleft}
\begin{subfigure}{\sharedwidth\textwidth}
  \centering
  \includegraphics[width=\linewidth, height = \sharedheight]{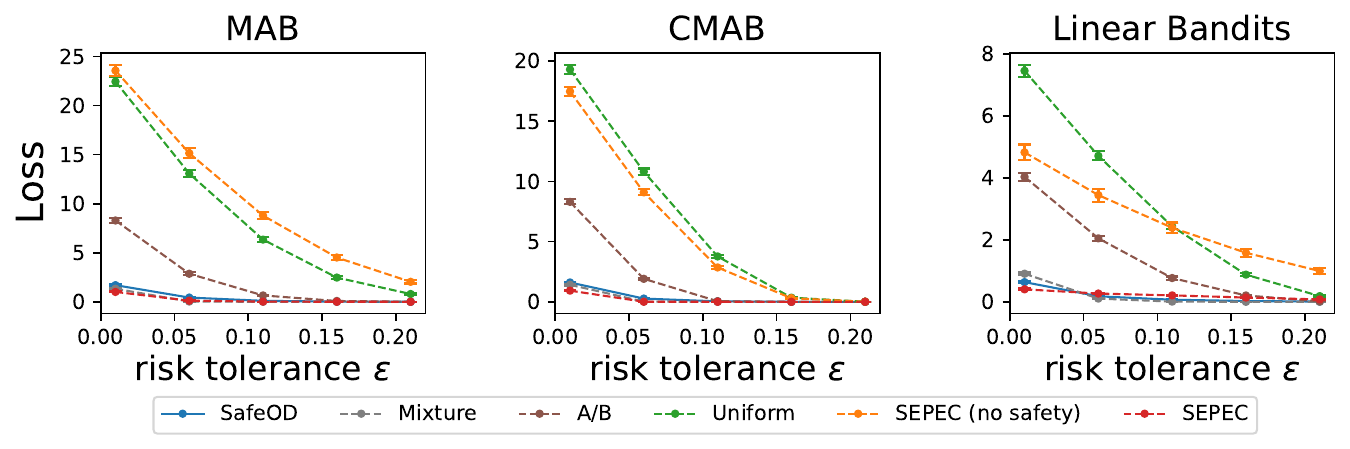}
    \vspace{\vbetween}
  \caption{Risk from safety violation.}
\end{subfigure}%

\hspace{\hleft}
\begin{subfigure}{\sharedwidth\textwidth}
  \centering
  \includegraphics[width=\linewidth, height = \sharedheight]{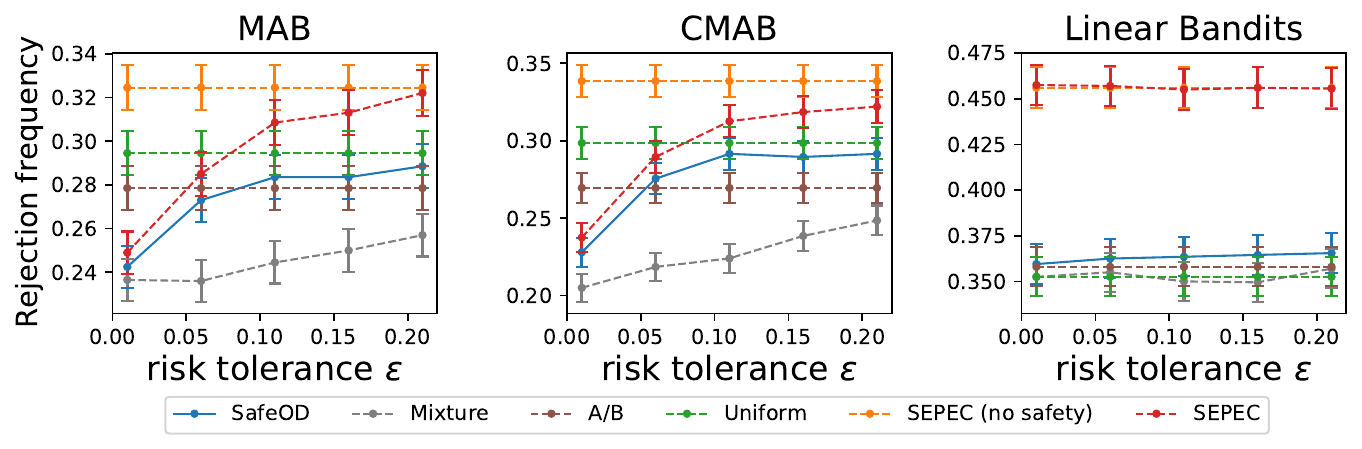}
  \vspace{\vbetween}
  \caption{Power of detecting $V(\pi_1) > V(\pi_0)$.}
\end{subfigure}%
\caption{Experiment results. 
The error bars indicate the standard errors of the averages.   
}\label{fig:simu}
\vspace{-0.5cm}
\end{figure}



For MAB, we set $K = 10$, $T=500$, $\delta = 0.05$, $\sigma_a \equiv 3$, sample $r_a \sim \text{Uniform}(0,1)$, and vary the risk tolerance $\epsilon$. 
We assume the noise is Gaussian, and construct $\mathcal{C}_\delta(\mD_0)$ from the posteriors of $\{r_a\}$, with $|\mD_0| = 100$ logged data points and the prior $\normal(r_a, 0.2^2)$ for every arm $a$. 
To mimic the real applications where risk and opportunity coexist, we generate $\pi_0$ and $\pi_1$ in the following manner: 
we first sample from $\text{Ber}(0.5)$ to determine if the null $H_0 : V(\pi_1) \le V(\pi_0)$ is true. 
If true, we sample $\Tilde{r}_a \sim \text{Uniform}(0,1)$, set $\pi_1$ as $\pi_1(a) \propto \Tilde{r}_a$, and then set $\pi_0$ as $\pi_0(a) \propto 0.5\Tilde{r}_a +  0.5 r_a$. 
If not, we set $\pi_0(a) \propto \Tilde{r}_a$ and $\pi_1(a) \propto 0.5\Tilde{r}_a +  0.5 r_a$.
The results are used to report the RMSEs and risks. 
We also run a test of level $5\%$ to detect $V(\pi_1) > V(\pi_0)$, and report the power when $H_1$ is true. 
The configurations for CMAB and linear bandits are similar. 
For CMAB, we consider $30$ contexts with  $\{p(\vx_i)\}$ sampled from $\text{Dirichlet}(\vone_{30})$. 
For linear bandits, we sample $\vthe^*$ from the standard multivariate normal and $\vx$ from the unit sphere uniformly, with $K = 100$, $d=5$, $T = 200$, and $|\mD_0| = 100$. 
A detailed description of experiments is in Appendix 
\ref{sec:more_expt_details}. 

\textbf{Results.} 
Results aggregated over $2000$ runs are reported in Figure \ref{fig:simu}. 
Overall, we observe that \name{SEPEC} consistently yields negligible loss from violating the safety constraint, and achieves lower RMSE and higher power compared with the other two safe policies (Mixture and SafeOD). 
The comparison with these two algorithms highlights the importance of carefully designing the exploration policy and utilizing task-specific structures. 
On the other hand, the A/B test and uniform exploration have slightly higher efficiency when the risk tolerance is very low, but are fairly risky. 
Moreover, compared with these two algorithms, the variant of \name{SEPEC} without safety constraints consistently shows better efficiency, which further supports the usefulness of a carefully designed exploration policy. 
As expected, when the safety constraint is less tight, the efficiency of \name{SEPEC} increases. 
The cost of satisfying the safety constraint is notably low in linear bandits, which is due to the generalization function.


To investigate the robustness of the findings, we repeat the experiment under a wide range of parameters in Appendix \ref{sec:more_expt}. 
In addition, to study the performance on real datasets, 
we conduct experiments using the MNIST dataset \cite{deng2012mnist} and present the results in Figure \ref{fig:real}, with more details given in Appendix \ref{sec:MNIST}. 
The findings are consistent. 


\begin{figure}[!t]
\hspace{-1cm}
\begin{subfigure}{0.48\textwidth}
  \centering
  \includegraphics[width=\linewidth, height = 3.2cm]{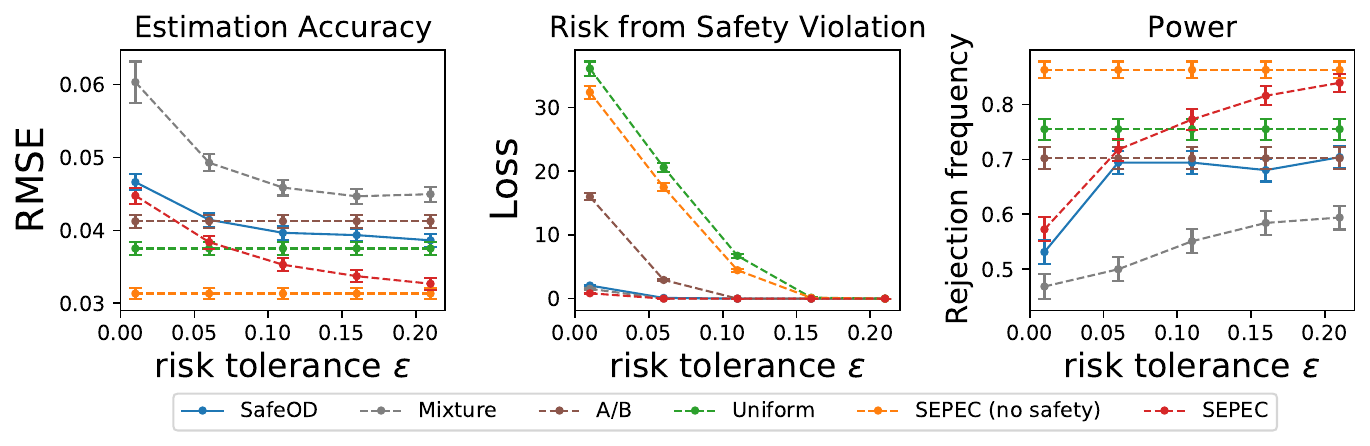}
  \caption{CMAB with IPW.}\label{fig:real_CMAB}
\end{subfigure}%
\hspace{.1cm}
\begin{subfigure}{0.48\textwidth}
  \centering
  \includegraphics[width=\linewidth, height = 3.2cm]{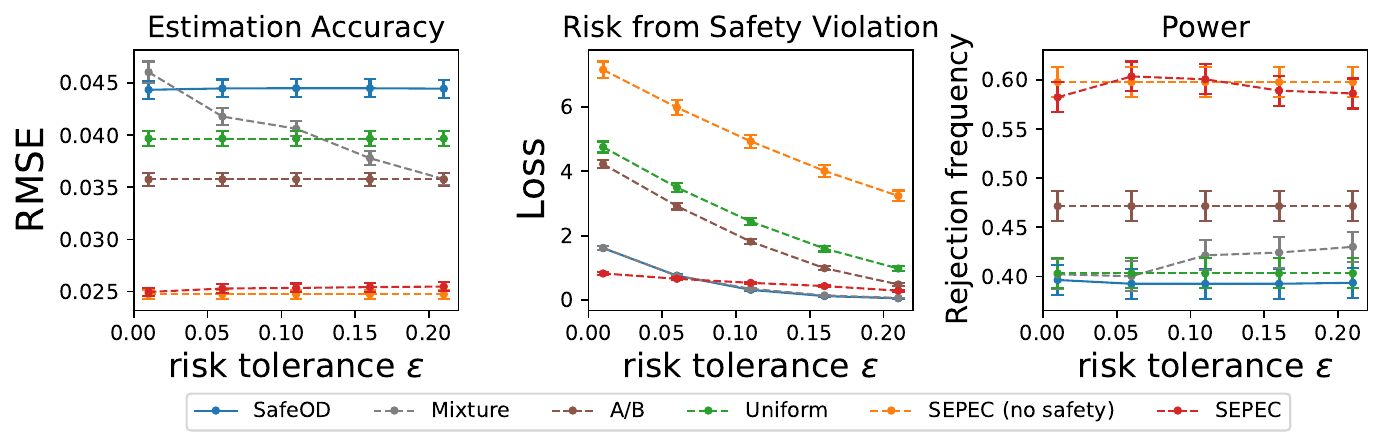}
  \caption{Linear bandits with DM.}\label{fig:real_LB}
\end{subfigure}%
\caption{Experiment results on the MNIST dataset. 
The errorbars indicate the standard errors of the averages.   
For methods that consider the safety constraint, the losses from safety violation are around zero and hence may overlap.}\label{fig:real}
\vspace{-0.4cm}
\end{figure}




\section{Related Work}

\textbf{Off-policy evaluation.}
The OPE literature can be roughly classified as 
the DM \cite{dudik2014doubly}, 
IPW \cite{horvitz1952generalization, li2015toward}, 
and doubly robust estimators \cite{dudik2014doubly, kallus2021optimal, su2020doubly}.
Few of the existing papers pay attention to data quality. 
\citet{oosterhuis2020taking} studies the exploration problem in a very specific ranking task, 
and only IPW is considered with no theoretical guarantee provided. 
\citet{tucker2022variance} is a parallel work that also studies data collection for OPE of bandit policies. 
The paper focuses on CMAB with IPW (and studies several extensions such as evaluating multiple policies), while we provide a more comprehensive study of various bandit setups and value estimators. 
Moreover, neither of the two papers studies the safety issue, which is practically important and introduces non-trivial challenges. 

\textbf{Conservative bandits.}
Safety constraints have been considered in the conservative bandits literature \cite{wu2016conservative, amani2019linear, moradipari2020stage}, where the key idea is to follow the baseline policy when the next taken action could be risky. 
This shares similar spirits with the mixture policy. 
Our objective of designing an exploration policy for OPE is orthogonal to this area which aims to minimize the cumulative regret. 

\textbf{Pure exploration.}
Similar to our work, the pure exploration literature  \cite{xu2018fully, degenne2019non, azizi2021fixed} aims to maximize the information gain, with the objective of  learning a good policy. 
Our work has a different objective in many aspects: 
we focus on the efficiency of policy evaluation, study non-adaptive policies, and further consider safety constraints.

\textbf{Optimal experiment design.}
Our problem is related to optimal experiment design \cite{allen2017near, baird2018optimal,  fontaine2021online}, where the goal is also to design a high-quality data collection rule for statistical inference. 
Only a few of these works consider the bandit problem and none of them study OPE. 
We also consider the safety constraint, which is of great practical importance and makes the problem more challenging. 


\vspace{-.1cm}
\section{Conclusions} 
This paper initiates the study of efficient exploration for bandit policy evaluation and additionally considers safety. 
We call this problem \name{SEPEC} and study several of its representative variants. 
At a high level, our work contributes to disentangling exploration and evaluation/optimization in bandits. 
We believe that this direction is of huge practical importance, 
as our exploration strategy is easy to deploy, safe, and enables better statistical inference.
Extensions to the Bayesian setup, sequential design, and reinforcement learning are all interesting next steps. 

In Section \ref{sec:contextual_MAB_IPW}, we assumed a finite context set for simplicity. 
A closer look at the derivations shows that, if there is no side information or we do not plan to use it (e.g., when we want worst-case safety guarantee), then this assumption is not required: 
every time when a new context is observed, 
we can just solve \eqref{eqn:CMAB_IPW_single} once. 
All nice properties are retained. 
In fact, we empirically observed that the efficiency loss is minimal, unless the risk budget is very low.  
The same argument applies when one does not need the safety constraint. 
The action set can be context-dependent as well. 
If one would like to use side information to guarantee high-probability safety with infinite contexts, then $\pi_e(a|\vx)$ must be parameterized (e.g., as a neural network). 
As a trade-off, additional assumptions on the policy class are needed. 
We leave differentiable policies for future research. 
Finally, considering a finite context set covers  applications where the context variables are discrete (e.g., day of the week, gender, age group, etc.). 


We also assumed in Section \ref{sec:contextual_MAB_IPW} that the context distribution is known. 
This is a common assumption \citep{wu2015algorithms, zhu2021safe} that simplifies theory. 
It is usually reasonable because we typically have a large set of historical contexts. 
If not, one can either
(i) optimize for every context independently as in \eqref{eqn:CMAB_IPW_single}, since possibly there is also not much side information; 
(ii) or replace $p(x)$ in the objective of \eqref{eqn:CMAB_IPW_2} by an upper bound and that in the constraint by a convex confidence region, for which we can similarly solve with a high-probability safety guarantee and worst-case efficiency guarantee.

\bibliographystyle{icml2022.bst}
\bibliography{0_MAIN_ICML}

\begin{thebibliography}{51}
\providecommand{\natexlab}[1]{#1}
\providecommand{\url}[1]{\texttt{#1}}
\expandafter\ifx\csname urlstyle\endcsname\relax
  \providecommand{\doi}[1]{doi: #1}\else
  \providecommand{\doi}{doi: \begingroup \urlstyle{rm}\Url}\fi

\bibitem[Abbasi-Yadkori et~al.(2011)Abbasi-Yadkori, P{\'a}l, and
  Szepesv{\'a}ri]{abbasi2011improved}
Abbasi-Yadkori, Y., P{\'a}l, D., and Szepesv{\'a}ri, C.
\newblock Improved algorithms for linear stochastic bandits.
\newblock \emph{Advances in neural information processing systems},
  24:\penalty0 2312--2320, 2011.

\bibitem[Allen-Zhu et~al.(2017)Allen-Zhu, Li, Singh, and Wang]{allen2017near}
Allen-Zhu, Z., Li, Y., Singh, A., and Wang, Y.
\newblock Near-optimal design of experiments via regret minimization.
\newblock In \emph{International Conference on Machine Learning}, pp.\
  126--135. PMLR, 2017.

\bibitem[Amani et~al.(2019)Amani, Alizadeh, and Thrampoulidis]{amani2019linear}
Amani, S., Alizadeh, M., and Thrampoulidis, C.
\newblock Linear stochastic bandits under safety constraints.
\newblock \emph{arXiv preprint arXiv:1908.05814}, 2019.

\bibitem[Azizi et~al.(2021)Azizi, Kveton, and Ghavamzadeh]{azizi2021fixed}
Azizi, M., Kveton, B., and Ghavamzadeh, M.
\newblock Fixed-budget best-arm identification in contextual bandits: A
  static-adaptive algorithm.
\newblock \emph{arXiv preprint arXiv:2106.04763}, 2021.

\bibitem[Baird et~al.(2018)Baird, Bohren, McIntosh, and
  {\"O}zler]{baird2018optimal}
Baird, S., Bohren, J.~A., McIntosh, C., and {\"O}zler, B.
\newblock Optimal design of experiments in the presence of interference.
\newblock \emph{Review of Economics and Statistics}, 100\penalty0 (5):\penalty0
  844--860, 2018.

\bibitem[Bertsimas \& Tsitsiklis(1997)Bertsimas and
  Tsitsiklis]{bertsimas1997introduction}
Bertsimas, D. and Tsitsiklis, J.~N.
\newblock \emph{Introduction to linear optimization}, volume~6.
\newblock Athena Scientific Belmont, MA, 1997.

\bibitem[Bottou et~al.(2013)Bottou, Peters, Qui{\~n}onero-Candela, Charles,
  Chickering, Portugaly, Ray, Simard, and Snelson]{bottou2013counterfactual}
Bottou, L., Peters, J., Qui{\~n}onero-Candela, J., Charles, D.~X., Chickering,
  D.~M., Portugaly, E., Ray, D., Simard, P., and Snelson, E.
\newblock Counterfactual reasoning and learning systems: The example of
  computational advertising.
\newblock \emph{Journal of Machine Learning Research}, 14\penalty0 (11), 2013.

\bibitem[Bouhtou et~al.(2010)Bouhtou, Gaubert, and
  Sagnol]{bouhtou2010submodularity}
Bouhtou, M., Gaubert, S., and Sagnol, G.
\newblock Submodularity and randomized rounding techniques for optimal
  experimental design.
\newblock \emph{Electronic Notes in Discrete Mathematics}, 36:\penalty0
  679--686, 2010.

\bibitem[Boyd \& Vandenberghe(2007)Boyd and Vandenberghe]{boyd2007localization}
Boyd, S. and Vandenberghe, L.
\newblock Localization and cutting-plane methods.
\newblock \emph{From Stanford EE 364b lecture notes}, 2007.

\bibitem[Boyd et~al.(2004)Boyd, Boyd, and Vandenberghe]{boyd2004convex}
Boyd, S., Boyd, S.~P., and Vandenberghe, L.
\newblock \emph{Convex optimization}.
\newblock Cambridge university press, 2004.

\bibitem[Brown \& Gajek(1990)Brown and Gajek]{brown1990information}
Brown, L.~D. and Gajek, L.
\newblock Information inequalities for the bayes risk.
\newblock \emph{The Annals of Statistics}, 18\penalty0 (4):\penalty0
  1578--1594, 1990.

\bibitem[Brumback(2009)]{brumback2009note}
Brumback, B.~A.
\newblock A note on using the estimated versus the known propensity score to
  estimate the average treatment effect.
\newblock \emph{Statistics \& Probability Letters}, 79\penalty0 (4):\penalty0
  537--542, 2009.

\bibitem[Cai et~al.(2020)Cai, Lu, and Song]{cai2020validation}
Cai, H., Lu, W., and Song, R.
\newblock On validation and planning of an optimal decision rule with
  application in healthcare studies.
\newblock In \emph{International Conference on Machine Learning}, pp.\
  1262--1270. PMLR, 2020.

\bibitem[Cai et~al.(2021)Cai, Shi, Song, and Lu]{cai2021deep}
Cai, H., Shi, C., Song, R., and Lu, W.
\newblock Deep jump learning for off-policy evaluation in continuous treatment
  settings.
\newblock \emph{Advances in Neural Information Processing Systems}, 34, 2021.

\bibitem[Degenne et~al.(2019)Degenne, Koolen, and M{\'e}nard]{degenne2019non}
Degenne, R., Koolen, W.~M., and M{\'e}nard, P.
\newblock Non-asymptotic pure exploration by solving games.
\newblock \emph{arXiv preprint arXiv:1906.10431}, 2019.

\bibitem[Deng(2012)]{deng2012mnist}
Deng, L.
\newblock The mnist database of handwritten digit images for machine learning
  research [best of the web].
\newblock \emph{IEEE Signal Processing Magazine}, 29\penalty0 (6):\penalty0
  141--142, 2012.

\bibitem[Dud{\'\i}k et~al.(2014)Dud{\'\i}k, Erhan, Langford, and
  Li]{dudik2014doubly}
Dud{\'\i}k, M., Erhan, D., Langford, J., and Li, L.
\newblock Doubly robust policy evaluation and optimization.
\newblock \emph{Statistical Science}, 29\penalty0 (4):\penalty0 485--511, 2014.

\bibitem[Fontaine et~al.(2021)Fontaine, Perrault, Valko, and
  Perchet]{fontaine2021online}
Fontaine, X., Perrault, P., Valko, M., and Perchet, V.
\newblock Online a-optimal design and active linear regression.
\newblock In \emph{International Conference on Machine Learning}, pp.\
  3374--3383. PMLR, 2021.

\bibitem[Frank et~al.(1956)Frank, Wolfe, et~al.]{frank1956algorithm}
Frank, M., Wolfe, P., et~al.
\newblock An algorithm for quadratic programming.
\newblock \emph{Naval research logistics quarterly}, 3\penalty0 (1-2):\penalty0
  95--110, 1956.

\bibitem[Horvitz \& Thompson(1952)Horvitz and
  Thompson]{horvitz1952generalization}
Horvitz, D.~G. and Thompson, D.~J.
\newblock A generalization of sampling without replacement from a finite
  universe.
\newblock \emph{Journal of the American statistical Association}, 47\penalty0
  (260):\penalty0 663--685, 1952.

\bibitem[Hsu et~al.(2011)Hsu, Kakade, and Zhang]{hsu2011analysis}
Hsu, D., Kakade, S.~M., and Zhang, T.
\newblock An analysis of random design linear regression.
\newblock \emph{arXiv preprint arXiv:1106.2363}, 2011.

\bibitem[Jaggi(2013)]{jaggi2013revisiting}
Jaggi, M.
\newblock Revisiting frank-wolfe: Projection-free sparse convex optimization.
\newblock In \emph{International Conference on Machine Learning}, pp.\
  427--435. PMLR, 2013.

\bibitem[Jiang \& Li(2016)Jiang and Li]{jiang2016doubly}
Jiang, N. and Li, L.
\newblock Doubly robust off-policy value evaluation for reinforcement learning.
\newblock In \emph{International Conference on Machine Learning}, pp.\
  652--661. PMLR, 2016.

\bibitem[Kallus et~al.(2021)Kallus, Saito, and Uehara]{kallus2021optimal}
Kallus, N., Saito, Y., and Uehara, M.
\newblock Optimal off-policy evaluation from multiple logging policies.
\newblock In \emph{International Conference on Machine Learning}, pp.\
  5247--5256. PMLR, 2021.

\bibitem[Kazerouni et~al.(2016)Kazerouni, Ghavamzadeh, Abbasi-Yadkori, and
  Van~Roy]{kazerouni2016conservative}
Kazerouni, A., Ghavamzadeh, M., Abbasi-Yadkori, Y., and Van~Roy, B.
\newblock Conservative contextual linear bandits.
\newblock \emph{arXiv preprint arXiv:1611.06426}, 2016.

\bibitem[Kveton et~al.(2021)Kveton, Konobeev, Zaheer, Hsu, Mladenov, Boutilier,
  and Szepesvari]{kveton2021meta}
Kveton, B., Konobeev, M., Zaheer, M., Hsu, C.-w., Mladenov, M., Boutilier, C.,
  and Szepesvari, C.
\newblock Meta-thompson sampling.
\newblock In \emph{International Conference on Machine Learning}, pp.\
  5884--5893. PMLR, 2021.

\bibitem[Lattimore \& Szepesv{\'a}ri(2020)Lattimore and
  Szepesv{\'a}ri]{lattimore2020bandit}
Lattimore, T. and Szepesv{\'a}ri, C.
\newblock \emph{Bandit algorithms}.
\newblock Cambridge University Press, 2020.

\bibitem[Lee(1988)]{lee1988constrained}
Lee, C. M.-S.
\newblock Constrained optimal designs.
\newblock \emph{Journal of Statistical Planning and Inference}, 18\penalty0
  (3):\penalty0 377--389, 1988.

\bibitem[Li et~al.(2011)Li, Chu, Langford, and Wang]{li2011unbiased}
Li, L., Chu, W., Langford, J., and Wang, X.
\newblock Unbiased offline evaluation of contextual-bandit-based news article
  recommendation algorithms.
\newblock In \emph{Proceedings of the fourth ACM international conference on
  Web search and data mining}, pp.\  297--306, 2011.

\bibitem[Li et~al.(2015)Li, Munos, and Szepesv{\'a}ri]{li2015toward}
Li, L., Munos, R., and Szepesv{\'a}ri, C.
\newblock Toward minimax off-policy value estimation.
\newblock In \emph{Artificial Intelligence and Statistics}, pp.\  608--616.
  PMLR, 2015.

\bibitem[Moradipari et~al.(2020)Moradipari, Thrampoulidis, and
  Alizadeh]{moradipari2020stage}
Moradipari, A., Thrampoulidis, C., and Alizadeh, M.
\newblock Stage-wise conservative linear bandits.
\newblock \emph{Advances in Neural Information Processing Systems}, 33, 2020.

\bibitem[Oosterhuis \& de~Rijke(2020)Oosterhuis and
  de~Rijke]{oosterhuis2020taking}
Oosterhuis, H. and de~Rijke, M.
\newblock Taking the counterfactual online: Efficient and unbiased online
  evaluation for ranking.
\newblock In \emph{Proceedings of the 2020 ACM SIGIR on International
  Conference on Theory of Information Retrieval}, pp.\  137--144, 2020.

\bibitem[Sachdeva et~al.(2020)Sachdeva, Su, and Joachims]{sachdeva2020off}
Sachdeva, N., Su, Y., and Joachims, T.
\newblock Off-policy bandits with deficient support.
\newblock In \emph{Proceedings of the 26th ACM SIGKDD International Conference
  on Knowledge Discovery \& Data Mining}, pp.\  965--975, 2020.

\bibitem[Shah \& Sinha(2012)Shah and Sinha]{shah2012theory}
Shah, K.~R. and Sinha, B.
\newblock \emph{Theory of optimal designs}, volume~54.
\newblock Springer Science \& Business Media, 2012.

\bibitem[Shi et~al.(2021)Shi, Wan, Chernozhukov, and Song]{shi2021deeply}
Shi, C., Wan, R., Chernozhukov, V., and Song, R.
\newblock Deeply-debiased off-policy interval estimation.
\newblock In \emph{International Conference on Machine Learning}, pp.\
  9580--9591. PMLR, 2021.

\bibitem[Slivkins(2019)]{slivkins2019introduction}
Slivkins, A.
\newblock Introduction to multi-armed bandits.
\newblock \emph{arXiv preprint arXiv:1904.07272}, 2019.

\bibitem[Su et~al.(2020)Su, Dimakopoulou, Krishnamurthy, and
  Dud{\'\i}k]{su2020doubly}
Su, Y., Dimakopoulou, M., Krishnamurthy, A., and Dud{\'\i}k, M.
\newblock Doubly robust off-policy evaluation with shrinkage.
\newblock In \emph{International Conference on Machine Learning}, pp.\
  9167--9176. PMLR, 2020.

\bibitem[Swaminathan et~al.(2016)Swaminathan, Krishnamurthy, Agarwal,
  Dud{\'\i}k, Langford, Jose, and Zitouni]{swaminathan2016off}
Swaminathan, A., Krishnamurthy, A., Agarwal, A., Dud{\'\i}k, M., Langford, J.,
  Jose, D., and Zitouni, I.
\newblock Off-policy evaluation for slate recommendation.
\newblock \emph{arXiv preprint arXiv:1605.04812}, 2016.

\bibitem[Thomas et~al.(2015)Thomas, Theocharous, and
  Ghavamzadeh]{thomas2015high}
Thomas, P., Theocharous, G., and Ghavamzadeh, M.
\newblock High confidence policy improvement.
\newblock In \emph{International Conference on Machine Learning}, pp.\
  2380--2388. PMLR, 2015.

\bibitem[Tran-The et~al.(2021)Tran-The, Gupta, Nguyen-Tang, Rana, and
  Venkatesh]{tran2021combining}
Tran-The, H., Gupta, S., Nguyen-Tang, T., Rana, S., and Venkatesh, S.
\newblock Combining online learning and offline learning for contextual bandits
  with deficient support.
\newblock \emph{arXiv preprint arXiv:2107.11533}, 2021.

\bibitem[Tsiatis(2007)]{tsiatis2007semiparametric}
Tsiatis, A.
\newblock \emph{Semiparametric theory and missing data}.
\newblock Springer Science \& Business Media, 2007.

\bibitem[Tucker \& Joachims(2022)Tucker and Joachims]{tucker2022variance}
Tucker, A.~D. and Joachims, T.
\newblock Variance-optimal augmentation logging for counterfactual evaluation
  in contextual bandits.
\newblock \emph{arXiv preprint arXiv:2202.01721}, 2022.

\bibitem[Vershynin(2010)]{vershynin2010introduction}
Vershynin, R.
\newblock Introduction to the non-asymptotic analysis of random matrices.
\newblock \emph{arXiv preprint arXiv:1011.3027}, 2010.

\bibitem[Wan et~al.(2021)Wan, Ge, and Song]{wan2021metadata}
Wan, R., Ge, L., and Song, R.
\newblock Metadata-based multi-task bandits with bayesian hierarchical models.
\newblock \emph{Advances in Neural Information Processing Systems}, 34, 2021.

\bibitem[Wang et~al.(2017)Wang, Agarwal, and Dud{\i}k]{wang2017optimal}
Wang, Y.-X., Agarwal, A., and Dud{\i}k, M.
\newblock Optimal and adaptive off-policy evaluation in contextual bandits.
\newblock In \emph{International Conference on Machine Learning}, pp.\
  3589--3597. PMLR, 2017.

\bibitem[Wu et~al.(2015)Wu, Srikant, Liu, and Jiang]{wu2015algorithms}
Wu, H., Srikant, R., Liu, X., and Jiang, C.
\newblock Algorithms with logarithmic or sublinear regret for constrained
  contextual bandits.
\newblock \emph{arXiv preprint arXiv:1504.06937}, 2015.

\bibitem[Wu et~al.(2016)Wu, Shariff, Lattimore, and
  Szepesv{\'a}ri]{wu2016conservative}
Wu, Y., Shariff, R., Lattimore, T., and Szepesv{\'a}ri, C.
\newblock Conservative bandits.
\newblock In \emph{International Conference on Machine Learning}, pp.\
  1254--1262. PMLR, 2016.

\bibitem[Xu et~al.(2018)Xu, Honda, and Sugiyama]{xu2018fully}
Xu, L., Honda, J., and Sugiyama, M.
\newblock A fully adaptive algorithm for pure exploration in linear bandits.
\newblock In \emph{International Conference on Artificial Intelligence and
  Statistics}, pp.\  843--851. PMLR, 2018.

\bibitem[Zanette et~al.(2021)Zanette, Dong, Lee, and
  Brunskill]{zanette2021design}
Zanette, A., Dong, K., Lee, J.~N., and Brunskill, E.
\newblock Design of experiments for stochastic contextual linear bandits.
\newblock \emph{Advances in Neural Information Processing Systems},
  34:\penalty0 22720--22731, 2021.

\bibitem[Zhou et~al.(2017)Zhou, Mayer-Hamblett, Khan, and
  Kosorok]{zhou2017residual}
Zhou, X., Mayer-Hamblett, N., Khan, U., and Kosorok, M.~R.
\newblock Residual weighted learning for estimating individualized treatment
  rules.
\newblock \emph{Journal of the American Statistical Association}, 112\penalty0
  (517):\penalty0 169--187, 2017.

\bibitem[Zhu \& Kveton(2022)Zhu and Kveton]{zhu2021safe}
Zhu, R. and Kveton, B.
\newblock Safe optimal design with applications in off-policy learning.
\newblock In \emph{Proceedings of the 25th International Conference on
  Artificial Intelligence and Statistics}, pp.\  2436--2447, 2022.

\end{thebibliography}

\clearpage

\onecolumn
\appendix
\newpage

\section{Additional Details of the Proposed Methods}
\subsection{Safety constraint with side information}\label{sec:appendix_constraint}
In this section, we give a few examples on how to construct the high-probability confidence (or credible) region $\mathcal{C}_\delta(\mD_0)$ and show why it is convex in these cases. 
For any vector $\vv$ and matrix $V$, we denote $||\vv||_V = \vv^T V^{-1} \vv$. 

For MAB, we can construct a confidence region by using an $(1-\delta / K)$-confidence interval for each arm separately. 
For example, suppose the noise is sub-Gaussian and there is at least one data point for each arm. 
We have $\prob \big[ |\hat{r}_a - r_a| / \sigma_a \ge \sqrt{2/n_a log(2/\delta)} \big] \le \delta$, 
where $n_a$ and $\hat{r}_a$ are the count and sample mean for arm $a$, respectively. 
Therefore, 
\begin{align*}
    \mathcal{C}_\delta(\mD_0) = \{\vr: max(\hat{r}_a - \sigma_a\sqrt{2/n_a log(2K/\delta)}, 0) \le r_a \le min(\hat{r}_a + \sigma_a\sqrt{2/n_a log(2K/\delta)}, 1), \forall a \in [K] \} 
\end{align*}
is a valid confidence region, based on the Bonferroni correction. 
We remark that, other choice is also possible than splitting the $\delta$ equally over arms. 

Alternatively, we can consider the Bayesian perspective. 
We additionally assume the existence of a prior over $r_a$ as $\normal(\mu_{a,0}, \sigma_{a,0}^2)$ for every arm $a$ independently. 
By assuming that the reward from arm $a$ follows $\normal(0, \sigma^2_a)$, 
we can derive $r_a | \mD_0 \sim \normal \big(
\hat{r}'_a, \hat{\sigma}^2_a\big)$, where $\hat{\sigma}^2_a = (1/\sigma_{a,0}^2 + n_a /\sigma^2_a)^{-1}$ and $\hat{r}'_a = \hat{\sigma}^2_a(\mu_{a,0} / \sigma_{a,0}^2 + n_a\hat{r}_a/\sigma^2_a)$. 
Therefore, one valid credible region is 
\begin{align*}
    \mathcal{C}_\delta(\mD_0) = \{\vr: max(\hat{r}'_a - z_{\delta/2K} \hat{\sigma}_{a}, 0) \le r_a \le min(\hat{r}'_a + z_{\delta/2K}\hat{\sigma}_{a}, 1), \forall a \in [K] \}, 
\end{align*}
where we construct an interval for each arm separately. 
Alternatively, utilizing the fact that these Gaussian variables are independent, we can construct a joint region. 

Notably, all examples above give us a convex set. 
The high-probability region in the CMAB case can be similarly derived. 

For linear bandits, based on $\mD_0$, with a given regularization parameter $\lambda$, we can construct a covariance matrix $\widehat{V}_0 = \sum_{(\vx_i, R_i) \in \mD_0} \vx_i \vx_i^T + \lambda \mathbf{I}$, an initial estimate $\hat{\vthe}_0 = \widehat{V}_0^-1 (\sum_{(\vx_i, R_i) \in \mD_0} \vx_i R_i)$, 
and a corresponding high-confidence region for $\vthe^*$ as $\mathcal{C}_\delta(\mD_0) = \{\vthe : ||\vthe - \hat{\vthe}_0||_{\widehat{V}_0} \le S_{\delta}\}$. 
By carefully choosing $S_{\delta}$ as given in Theorem 2 of \citet{abbasi2011improved}, we can guarantee that $\prob\big(\vthe^* \in \mathcal{C}_\delta(\mD_0)\big) \ge 1-\delta$. 

For the Bayesian perspective, we additionally assume the existence of a prior over $\vthe$ as 
$\normal\big(\vmu_{\vthe}, \Sigma_{\vthe}\big)$, which can be learned either from domain knowledge or via meta-learning \cite{kveton2021meta, wan2021metadata}. 
Assume that the noise follows $\normal(0, \sigma^2)$. 
The standard Bayesian results on Gaussian linear regression tell us that 
\begin{align*}
    \vthe | \mD_0 \sim \normal\Big( \widehat{\Sigma} \big(\Sigma_{\vthe}^{-1}\vmu_{\vthe}  + \sum_{(\vx_i, R_i) \in \mD_0} \vx_i R_i / \sigma^2 \big) , \widehat{\Sigma}  \Big), 
    \widehat{\Sigma} = \big(\Sigma_{\vthe}^{-1} + \sum_{(\vx_i, R_i) \in \mD_0} \vx_i \vx_i^T / \sigma^2 \big)^{-1}
\end{align*}
Let $\hat{\vthe} =  \widehat{\Sigma} \big(\Sigma_{\vthe}^{-1}\vmu_{\vthe}  + \sum_{(\vx_i, R_i) \in \mD_0} \vx_i R_i / \sigma^2 \big)$. 
We can then define  the ellipsoid as 
$\mathcal{C}_\delta(\mD_0) = \{\vthe : ||\vthe - \hat{\vthe}||_{\widehat{V}} \le \chi^2_{d}(1-\delta)\}$.


\subsection{Connections with hypothesis testing}\label{sec:other_testing}
For linear bandits, the problem reduces to the random design analysis of linear models   \cite{hsu2011analysis}. 
We assume homoscedasticity, i.e., $\var(\epsilon_t) \equiv \sigma^2$. 
We focus on that $\mD_{0}$ is empty for simplicity, and it is straightforward to extend to the case with $\mD_{0}$. 
By the central limit theorem and the Slutsky’s lemma, 
we can establish that 
\begin{align*}
    \sqrt{T}(\hat{\vbeta} - \vbeta) 
     &\overset{d}{\to} \normal\Big(\vzero,  
     \sigma^2
     \big(
     \sum_\vx \pi_e(\vx) \vx \vx^T
     \big)^{-1}
     \Big), 
     T \to \infty. 
\end{align*}
which implies 
\begin{align*}
    \sqrt{T}
    \Big[
    \big(\widehat{V}_{DM}(\pi_1; \mD) - \widehat{V}_{DM}(\pi_0; \mD) \big)
    - \big(V(\pi_1; \mD) - V(\pi_0; \mD) \big)
    \Big]
     &\overset{d}{\to} \normal\Big(\vzero, 
     \sigma^2 
    \bar{\vphi}^T_{\Delta \pi}  \big(\sum_\vx \pi_e(\vx) \vx \vx^T\big)^{-1} \bar{\vphi}_{\Delta \pi}
     \Big), 
     T \to \infty. 
\end{align*}
Therefore, by similar derivations as in the MAB case, we can design a valid testing procedure and identify that the testing power is determined by $\bar{\vphi}^T_{\Delta \pi} \big( \sum_\vx \pi_e(\vx) \vx \vx^T\big)^{-1} \bar{\vphi}_{\Delta \pi}$, i.e., our optimization objective. 

Regarding CMAB with IPW, note the fact the the estimator, as in the MAB case, is also an U-statistic (i.e., the average of i.i.d samples). 
Therefore, we can similar derive its asymptotic distribution and relate the asymptotic variance (and hence the power) with our optimization objective. 
For example, 
let $\sigma(\pi_e) {=} \sqrt{\var(\pi_e(A_t|\vx_t)^{-1}\pi_{\Delta}(A_t|\vx_t)R_t)}$ and $\hat{\sigma}(\pi_e)$ be a consistent estimator (typically the plug-in estimator with sample variance). 
By the central-limit theorem and the Slutsky’s lemma, we again have  
\begin{align*}
\frac{\sqrt{T}}{\hat{\sigma}(\pi_e)}
    \Big[
    \big(\widehat{V}(\pi_1; \mD_{e}) 
    {-} \widehat{V}(\pi_0; \mD_{e}) \big)
    {-} \big(V(\pi_1) {-} V(\pi_0)\big)\Big]
    \overset{d}{\to} \normal(0, 1), 
\end{align*}
when $T$ grows.  
The remaining discussions are exactly the same with the MAB case.

\subsection{Optimization algorithm for MAB and CMAB with IPW}\label{sec:optimization_MAB}
As we mentioned in the main text, at the first glance, it is challenging to solve 
\eqref{eqn:MAB_IPW_w_logged} as it involves infinite number of linear constraints. 
Fortunately, both the objective and the feasible set are convex, and so we can adapt the cutting-plane method \cite{boyd2007localization} to solve.

As an iterative algorithm, there are two main steps in the cutting-plane method. 
At every iteration, we need to 
(i) proposed a \textit{query point} $\pi_e^{(i)}$, and then 
(ii) check with an oracle on whether or not $\pi_e^{(i)}$ is in a target convex set (the feasible set in our case): 
if true, then we can terminate; otherwise, we find a plane to separate $\pi_e^{(i)}$ and the target convex set (i.e., the cutting plane).

The choices of $\pi_e^{(i)}$ and the cutting plane are both central to the computational efficiency. 
Following  \cite{zhu2021safe}, we design the algorithm as in Algorithm \ref{alg:CP_MAB}, where both are obtained by a solving a simple subproblem. 
Specifically, the one for $\pi_e^{(i)}$ has a convex objective function and a set of linear constraints, and the one for $\vr^{(i)}$ is a linear programming problem. 
The optimization algorithm for the CMAB case can be designed similarly. 
The convergence analysis is provided in \citet{boyd2007localization}.

\begin{algorithm}[!h]
\SetAlgoLined
\KwData{
$\pi_0$, $\pi_1$
, $\mathcal{C}_\delta(\mD_0)$
, $\epsilon$
}

Set $i = 0$, $\pi_e^{(0)} = \pi_0$ and $\Theta = \varnothing$. 

\While{Not converge}{
    i = i + 1
    
    Apply convex optimization algorithms to obtain $\pi_e^{(i)}$ by solving
    \begin{equation*}
        \begin{split}
            &\argmin_{\pi_e \in \Pi_{\Delta}} \sum_{a \in [K]} \frac{\pi^2_{\Delta}(a)}{\pi_e(a)} \\
            &s.t. \;\;  \vr^T(\pi_e - (1-\epsilon) \pi_0) \ge 0, \forall \vr \in \Theta. 
        \end{split}
    \end{equation*}
    
    \uIf{$min_{\vr \in\mathcal{C}_\delta(\mD_0)} \vr^T(\pi_e^{(i)} - (1-\epsilon) \pi_0) \ge  0$}{
    Terminate the iteration
    }
    \Else{
    Solve $\vr^{(i)} = \argmin_{\vr \in\mathcal{C}_\delta(\mD_0)} \vr^T(\pi_e^{(i)} - (1-\epsilon) \pi_0)$  \tcp*{Constraint is maximally violated}
    
    Update $\Theta = \Theta \cup \{\vr^{(i)}\}$
  }
}

\textbf{Output:} $\pi_e = \pi_e^{(i)}$

 \caption{Safe MAB exploration with the cutting-plane method}\label{alg:CP_MAB}
\end{algorithm}


\subsection{Optimization algorithm for linear bandits with DM}\label{sec:optimization_LB}
We first introduce how to solve for a finite set of constraints $\Theta$: 
\begin{equation*}
    \begin{split}
&\argmin_{\pi_e \in \Delta_{K-1}} \;
\mJ(\pi_e) = 
\bar{\vphi}^T_{\Delta \pi}
    (T\sum_{\vx \in \mathcal{A}} \pi_e(\vx) \vx \vx^T+ \vPhi_0^{T}\vPhi_0)^{+}
    \bar{\vphi}_{\Delta \pi}\\
&s.t. \; \max_{\vthe \in \Theta} \big((1-\epsilon) \pi_0 - \pi_e\big)^T \mX \vthe \le  0. 
    \end{split}
\end{equation*}
We propose to adapt the Frank–Wolfe (FW) algorithm \citep{frank1956algorithm}, which is a projection-free algorithm for convex optimization and it enjoys nice convergence property  \citep{frank1956algorithm, lattimore2020bandit, jaggi2013revisiting}. 
We summarize in Algorithm \ref{alg:FW}. 
For Step 1, we note that 
the partial derivative can be derived as
\begin{equation*}
    \begin{split}
\frac{\partial \mJ(\pi_e)}{\partial \pi_e(\vx_i)}
&= 
-T
\Big\{
\bar{\vphi}^T_{\Delta \pi}
(T\vG(\pi_e) + \vG_0)^{-1} \vx_i \Big\}^2
    \end{split}
\end{equation*}
where $\vG_0 = \vPhi_0^T \vPhi_0$ and $\vG(\pi_e) = \sum_\vx \pi_e(\vx) \vx \vx^T$. 
For Step 2, we notice that it is a linear programming problem.


\begin{algorithm}[!h]
\SetAlgoLined
\KwData{
$\pi_0$
, $\mathcal{C}_\delta(\mD_0)$
, $\epsilon$
, $\mX$
, $\vPhi_0$
, $\bar{\vphi}_{\Delta \pi}$
}

Set $i = 0$, $\pi_e^{(0)} = \pi_0$ and $\Theta = \varnothing$. 

\While{TRUE}{
    i = i + 1
    
    1. Compute the gradient $d^{(i)} = \partial \mJ(\pi_e)/\partial \pi_e|_{\pi_e=\pi_e^{(i)}}$
    
    2. Compute a feasible direction as $\Delta \pi_e$ by solving 
    \begin{equation*}
    \begin{split}
    &\argmin_{w \in \Delta_{K-1}} \;
    d^{(i)} w\\
    &s.t. \; \max_{\vthe \in \Theta} \big((1-\epsilon) \pi_0 - w\big)^T \mX \vthe \le  0. 
    \end{split}
    \end{equation*}
    
    3. Run line search to find the optimal step size $\lambda^{(i)}$ by solving
    $
    \argmin_{\lambda \in [0,1]} \mJ\Big(\pi_e^{(i)} + \lambda (\Delta \pi_e - \pi_e^{(i)}) \Big)
    $

    4. Update $\pi_e^{(i+1)} = \pi_e^{(i)} + \lambda^{(i)} (\Delta \pi_e - \pi_e^{(i)}) $
}
\textbf{Output:} $\pi_e = \pi_e^{(i)}$
 \caption{Optimization for linear bandits with the FW algorithm}\label{alg:FW}
\end{algorithm}



\begin{algorithm}[!h]
\SetAlgoLined
\KwData{
$\pi_0$
, $\mathcal{C}_\delta(\mD_0)$
, $\epsilon$
, $\mX$
, $\vPhi_0$
, $\bar{\vphi}_{\Delta \pi}$
}

Set $i = 0$, $\pi_e^{(0)} = \pi_0$ and $\Theta = \varnothing$. 

\While{TRUE}{
    i = i + 1

    Run the FW algorithm to obtain $\pi_e^{(i)}$ by solving
    \begin{align*}
&\argmin_{\pi_e \in \Delta_{K-1}} \;
\mJ(\pi_e) = 
\bar{\vphi}^T_{\Delta \pi}
    (T\sum_\vx \pi_e(\vx) \vx \vx^T+ \vPhi_0^{T}\vPhi_0)^{-1}
    \bar{\vphi}_{\Delta \pi}\\
&s.t. \;  \big((1-\epsilon) \pi_0 - \pi_e\big)^T \mX \vthe \le  0, \forall \vthe \in \Theta
    \end{align*}
    
    \uIf{$max_{\vthe \in C_{\delta}} \big((1-\epsilon) \pi_0 - \pi_e^{(i)}\big)^T \mX \vthe \le  0$}{
    Terminate the iteration
    }
    \Else{
    Solve $\vthe^{(i)} = \argmax_{\vthe \in C_{\delta}} \big((1-\epsilon) \pi_0 - \pi_e^{(i)} \big)^T \mX \vthe$  \tcp*{Constraint is maximally violated}
    
    Update $\Theta = \Theta \cup \{\vthe^{(i)}\}$
  }
    
    
}

\textbf{Output:} $\pi_e = \pi_e^{(i)}$

 \caption{Safe linear bandits exploration with the FW algorithm and the cutting-plane method}\label{alg:FW_CP}
\end{algorithm}




    
    
    
    



Next, we integrate the cutting-plane method and the FW algorithm to solve the most challenging problem \eqref{eqn:nonstochastic linear bandit with side info}, where there exists a infinite number of constraints. 
Refer to Appendix \ref{sec:optimization_MAB} for an introduction to the cutting-plane method. 
We note that, since the cutting-plane method only requires a feasible query point in every iteration instead of running the whole FW algorithm every time, one can stop the FW after a few iterations. 
This does not affect the convergence over the whole algorithm. 

In Algorithm \ref{alg:FW_CP}, a key step is to find a good cutting plane, or equivalently, $\vthe \in C_\delta$ on which the constraint is maximally violated by the current policy $\pi^{(i)}$.
Assume $\mathcal{C}_\delta(\mD_0) = \{\vthe : ||\vthe - \hat{\vthe}_0||_{\widehat{V}_0} \le S_{\delta}\}$ for some $\hat{\vthe}_0$ and $\widehat{V}_0$. 
Let $\vv_{\pi^{(i)}} = \Big[\big((1-\epsilon) \pi_0 - \pi^{(i)}\big)^T \mX \Big]^T$. 
We need to maximize a linear function on an ellipsoid. 
Fortunately, the solution can be obtained explicitly \citep{zhu2021safe} as 
\begin{align*}
\vthe^{(i)} = 
    \argmax_{\vthe \in \mathcal{C}_\delta(\mD_0)} 
\Big[
\big((1-\epsilon) \pi_0 - \pi^{(i)}\big)^T \mX 
\Big]
\vthe 
= 
\hat{\vthe}_0
+ 
\frac{\widehat{V}_0 \vv_{\pi^{(i)}}}{\sqrt{\vv_{\pi^{(i)}}^T \widehat{V}_0 \vv_{\pi^{(i)}} /  S_\delta}}.  
\end{align*}


\section{Extensions}\label{sec:other_setup}
\subsection{MAB with DM}\label{sec:MAB_DM}
An alternative estimator for MAB is the direct method (DM). 
In MAB, DM essentially plugs-in the estimated mean of every arm as 
$\widehat{V}(\pi; \mD) = \sum_a \Big[\pi(a) \frac{\sum_{(A_t, R_t) \in \mD} R_t \I(A_t = a)}{\sum_{(A_t, R_t) \in \mD} \I(A_t = a)} \Big]$. 
We note this estimator is actually equivalent to 
$|\mD|^{-1}\sum_{(A_i, R_i) \in \mD} \frac{\pi(A_i)}{\hat{\pi}_e^*(A_i)}R_i$, where $\hat{\pi}_e^*(a) = |\mD|^{-1}\sum_{(A_i, R_i) \in \mD} \I(A_t = a)$, i.e., the IPW-form estimator with the estimated propensity $\hat{\pi}_e^*(a)$. 
A well-known but perhaps counter-intuitive result \cite{brumback2009note} is that, even if $\mD$ is collected by following $\pi_e$ and $\pi_e$ is known, $\widehat{V}_{IPW}(\pi; \mD)$ is still more efficient than plugging-in the known function $\pi_e$. 

The following result is standard \cite{li2015toward} and we just recap here for completeness: 
\begin{align*}
    &\widehat{V}(\pi; \mD) - V(\pi)\\
    &=\sum_a \Big[\pi(a) \frac{\sum_{(A_t, R_t) \in \mD} R_t \I(A_t = a)}{\sum_{(A_t, R_t) \in \mD} \I(A_t = a)} \Big]
    - V(\pi)\\
   &=\sum_a \Big[\pi(a) \big( \frac{\sum_{(A_t, R_t) \in \mD} R_t \I(A_t = a)}{\sum_{(A_t, R_t) \in \mD} \I(A_t = a)}
   - r_a \big)
   \Big]\\
   &=\sum_a \Big[\pi(a) \big( \frac{\sum_{(A_t, R_t) \in \mD} (R_t- r_a) \I(A_t = a)}{\sum_{(A_t, R_t) \in \mD} \I(A_t = a)}
    \big)
   \Big]\\
   &=\sum_a \Big[\pi(a) \big( \frac{ |\mD|^{-1} \sum_{(A_t, R_t) \in \mD} (R_t- r_a) \I(A_t = a)}{|\mD|^{-1} \sum_{(A_t, R_t) \in \mD} \I(A_t = a)}
    \big)
   \Big]\\
   &\convergeD \normal\Big(0, \sum_a \frac{\pi^2(a)}{\pi_e(a)}\sigma_a^2 \Big). 
\end{align*}
By the additivity, we can see this relationship holds for the value difference $V(\pi_1)-V(\pi_0)$ as well. 
All of our discussions on the safety constraints and optimization algorithms can be directly extended to this case. 
Finally, similar extensions can be constructed for CMAB as well.

\subsection{Stochastic contextual linear bandits}\label{sec:contextual_LB}
Next, we consider the contextual bandits setting with a linear generalization function, i.e., the stochastic contextual linear bandits problem. 
At every round $t$, a stochastic context $\vx_t$ will be sampled i.i.d. and can not be known \textit{a priori}.  
The agent will then choose an arm $A_t$, which together with $\vx_t$ gives us the transformed feature vector $\vphi_t = \vphi_{\vx_t, A_t}$. 
Let $\bar{\vphi}_{\pi} = \Mean_{\vx \sim p(\vx), a \sim \pi(a \mid \vx) }   \vphi_{\vx,a}$. 
Assume $p(\vx)$ is known and hence $\bar{\vphi}_{\pi}$ is also known. 
Given an estimate $\hat{\vthe}$ obtained via least-square regression, 
recall our value estimator via DM is $\widehat{V}_{\hat{\vthe}}(\pi) = \Mean_{\vx \sim p(\vx), a \sim \pi(a \mid \vx)} \vphi^T_{\vx,a}  \hat{\vthe}  
= \big[ \Mean_{\vx \sim p(\vx), a \sim \pi(a \mid \vx)} \vphi^T_{\vx,a}\big]  \hat{\vthe} 
= \bar{\vphi}^T_{\pi} \hat{\vthe}$. 
Let $\bar{\vphi}_{\Delta\pi} = \bar{\vphi}_{\pi_1} - \bar{\vphi}_{\pi_0}$. 
W.l.o.g.,  we assume the noise variance is upper bounded by $1$. 
Without other information or constraints, 
by similar arguments as in Section \ref{sec:Nonstochastic linear bandits}, 
our objective can be transformed as follows
\begin{equation}
    \begin{split}
    &\var(\widehat{V}(\pi_1; \mD_{e} \cup \mD_{0}) - \widehat{V}(\pi_0; \mD_{e} \cup \mD_{0})) \\
    &= \bar{\vphi}^T_{\Delta\pi} cov(\hat{\vthe})  \bar{\vphi}_{\Delta\pi}\\
    &= 
    \bar{\vphi}^T_{\Delta\pi}
    \Big\{
    \Mean_{\vPhi \sim \pi_e, \vx} \Big[ cov(\hat{\vthe} \mid \vPhi) \Big]
    + cov_{\vPhi \sim \pi_e, \vx} \Big[ \Mean(\hat{\vthe} \mid \vPhi) \Big]
    \Big\}
    \bar{\vphi}_{\Delta\pi}\\
    &= 
    \bar{\vphi}^T_{\Delta\pi}
    \Big\{
\Mean_{\vPhi \sim \pi_e, \vx} \Big[ (\vPhi^T\vPhi + \vPhi_0^T\vPhi_0)^{-1}\Big]\Big\}
    \bar{\vphi}_{\Delta\pi}\\
    \end{split}
\end{equation}
Here, $\vPhi$ and $\vPhi_0$ is the feature matrix stacked over points in $\mD$ and $\mD_0$, respectively. 
More specifically, $\vPhi = (\vphi_{\vx_1, A_1}, \dots, \vphi_{\vx_T, A_T})^T$. 
We use $\vPhi \sim \pi_e, \vx$ to emphasize the expectation is taken over $\vPhi$, which depends on both the policy $\pi_e$ and the  stochastic context $\vx$.


Besides, the safety constraint also needs to be satisfied. 
Therefore, our problem can be written as 
\begin{equation}
    \begin{split}
    &\argmin_{\pi_e}
    \bar{\vphi}^T_{\Delta\pi}
    \Big\{
\Mean_{\vPhi \sim \pi_e, \vx} \Big[ (\vPhi^T\vPhi + \vPhi_0^T\vPhi_0)^{-1}\Big]\Big\}
    \bar{\vphi}_{\Delta\pi}\\
    &s.t. \; 
    \big[
    \bar{\vphi}_{\pi_e^*}^T - (1-\epsilon) \bar{\vphi}_{\pi_0}^T
    \big]\vthe
    \ge 0, \forall \vthe \in \mathcal{C}_\delta(\mD_0), \\
    &\quad \; 
    \pi_e(\cdot \mid \vx) \in \Delta_{K - 1}, \forall \vx \in \mathcal{X}
    \end{split}
\end{equation}
The challenges on optimization include that (i) both the objective and $\bar{\vphi}_{\pi_e}$ in the constraint can have complex dependency on $\pi_e$, 
(ii) the constraint implicitly includes infinite linear constraints, 
and (iii) the stochasticity in the context is hard to handle, unlike in Section \ref{sec:Nonstochastic linear bandits}. 

When there is a finite number of contexts, 
following similar arguments as in Section \ref{sec:Nonstochastic linear bandits}, 
we can consider the following surrogate objective instead, which can be similarly proved as the asymptotic variance of our procedure: 
\begin{equation}
    \begin{split}
            &\argmin_{\pi_e} 
    \bar{\vphi}^T_{\Delta\pi}
    \Big\{ 
    \Big[ (
    T \sum_{\vx, a} p(\vx) \pi_e(a | \vx) \vphi_{\vx, a} \vphi^T_{\vx, a}
     + \vPhi_0^T\vPhi_0)^{-1}\Big]\Big\}
    \bar{\vphi}_{\Delta\pi}\\
    &s.t. \; \min_{\vthe \in \mathcal{C}_\delta(\mD_0)} \quad 
    \big[
    \sum_{\vx, a} p(\vx) \pi_e(a | \vx) \vphi_{\vx, a}
    - (1-\epsilon) \bar{\vphi}_{\pi_0}
    \big]^T\vthe
    \ge 0. 
    \end{split}
\end{equation}
The problem can be solved similarly as in \eqref{eqn:nonstochastic linear bandit with side info} by utilizing the convexity. 
See Appendix  \ref{sec:optimization_LB} for details.

\subsection{Pseudo-inverse estimator for linear bandits and its contextual version}\label{sec:PI}
Under the linear generalization assumption, besides the DM estimator, we can also consider the Pseudo-Inverse (PI) estimator \cite{swaminathan2016off}. 
Instead of inferring $\vthe$, PI directly constructs a weighted average of the observed rewards as in IPW, but it also utilizes the linear structure. 
PI is particular useful when being applied to some structured problem such as slate recommendation  \cite{swaminathan2016off}. 
Let $\vG(\pi) = \sum_\vx \pi(\vx) \vx \vx^T$. 
We first note that 
\begin{align*}
    V(\pi; \mD) 
    = \bar{\vphi}^T_{\pi} \vthe
    = \bar{\vphi}^T_{\pi} \vG(\pi)^{-1}  \sum_{\vx} \pi(\vx) \vx (\vx^T\vthe)
\end{align*}
Motivated by this relationship, the PI estimator is constructed as 
\begin{align}
    \widehat{V}_{PI}(\pi; \mD) 
    = |\mD|^{-1} \sum_{(\vx_i, R_i) \in \mD} \bar{\vphi}^T_{\pi} \vG(\pi)^{-1} (R_i \vx_i)
    = |\mD|^{-1} \bar{\vphi}^T_{\pi} \vG(\pi)^{-1}  \sum_{(\vx_i, R_i) \in \mD}R_i \vx_i
    \label{eqn:PI_estimator}
\end{align}
Let $\vPhi$ be the feature matrix stacked over points in $\mD$. 
We first derive the finite-sample conditional variance $\var(\widehat{V}_{PI}(\pi_1; \mD) | \vPhi)$ as 
\begin{align*}
    \var(\widehat{V}_{PI}(\pi_1; \mD) | \vPhi)
    = |\mD|^{-2}\sum_{(\vx_i, R_i) \in \mD} \Big[\bar{\vphi}^T_{\pi_1} \vG(\pi_1)^{-1}\vx_i \Big]^2  \var(R_i \mid \vx_i)
\end{align*}
For simplicity of notations, w.l.o.g.,  we assume the noise variance is $1$. 
By the law of total variance, we can obtain that
\begin{equation}
    \begin{split}
    &T \times \var(\widehat{V}_{PI}(\pi_1; \mD_e))\\
    &=T \times  \Mean_{\vPhi \sim \pi_e} \var(\widehat{V}_{PI}(\pi_1; \mD_e) | \vPhi)
    +T \times  \var_{\vPhi \sim \pi_e} \Mean(\widehat{V}_{PI}(\pi_1; \mD_e)|\vPhi)\\
    &= \Mean_{\vx_i \sim \pi_e}  \Big[\bar{\vphi}^T_{\pi_1} \vG(\pi_1)^{-1}\vx_i \Big]^2 
    + T \times \var_{\vPhi \sim \pi_e} \Big\{ T^{-1} \bar{\vphi}^T_{\pi} \vG(\pi)^{-1}  \vPhi^T \vPhi \vthe \Big\}\\
    &\to  \bar{\vphi}^T_{\pi_1} \vG(\pi_1)^{-1} \vG(\pi_e) \vG(\pi_1)^{-1} \bar{\vphi}_{\pi_1}
    + \var_{\vx \sim \pi_e} \Big\{ \bar{\vphi}^T_{\pi} \vG(\pi)^{-1}  \vx \vx^T \vthe \Big\}.  
    \end{split}
\label{eqn:raw_PI}
\end{equation}
The proof for the asymptotic variance statement (i.e., the last row) is by similar arguments as in Appendix \ref{sec:proof_LB}.

By similar arguments as in the IPW for MAB case, 
the raw objective \eqref{eqn:raw_PI} is infeasible to solve, as it involves the unknown parameter $\vthe$. 
Therefore, we relax with the upper bound of the unknown parameters and focus on minimizing  the \textit{intrinsic uncertainty} term from the reward noise, i.e., 
$\bar{\vphi}^T_{\pi_1} \vG(\pi_1)^{-1} \vG(\pi_e) \vG(\pi_1)^{-1} \bar{\vphi}_{\pi_1}$. 
Notice that, under the linear generalization assumption, all discussions in Section \ref{sec:Nonstochastic linear bandits} regarding the safety constraint still apply here. 
Therefore, this problem can be similarly solved as in Section \ref{sec:Nonstochastic linear bandits}, via combining the cutting-plane method with the FW algorithm. 


Besides, 
by noting that \eqref{eqn:PI_estimator} is linear in $\bar{\vphi}^T_{\pi} \vG(\pi)^{-1}$, it is straightforward to extend to studying the value difference $V(\pi_1) - V(\pi_0)$. 
Moreover, as in \citet{swaminathan2016off} and \citet{zhu2021safe}, the PI estimator and the discussions above can be extended to the contextual setup in a straightforward manner. 



\subsection{Doubly robust estimator for MAB and CMAB}\label{sec:DR}

Besides IPW and DM, the doubly robust (DR) estimator is another popular OPE estimator, both in bandits \citep{dudik2014doubly} and in reinforcement learning \citep{shi2021deeply}. 
We analyze the DR estimator in the MAB setup, and the CMAB case is similar. 
Let $\widehat{V}_{DM}$ be a consistent DM-type estimator. 
The DR value estimator \cite{tsiatis2007semiparametric} is define as 
\begin{align*}
    \widehat{V}_{DR}(\pi_1; \mD_{e}) = \sum_t \frac{\pi_{1}(A_t)}{\pi_e(A_t)} (R_t - \widehat{V}_{DM}(\pi_1; \mD_{e})) + \widehat{V}_{DM}(\pi_1; \mD_{e}). 
\end{align*}
Due to the complex structure, it is well-known that its finite-sample variance does not yield a tractable form as IPW and DM do, and therefore people commonly focus on its asymptotic variance \cite{tsiatis2007semiparametric} 
\begin{align*}
&T \times \text{AsmypVar}\Big(\widehat{V}_{DR}(\pi_1; \mD_{e})) \Big)\nonumber\\
&= 
\sum_{a \in [K]} \frac{\pi^2_{1}(a)}{\pi_e(a)} \sigma^2_a 
+ \var_{A_t \sim \pi_e}\Big[\frac{\pi_{1}(A_t)}{\pi_e(A_t)}\big(r_{A_t} - V(\pi_1) \big)\Big]\\
 &= 
\sum_{a \in [K]} \frac{\pi^2_{1}(a)}{\pi_e(a)} \sigma^2_a 
+ \sum_{a \in [K]} \frac{\pi^2_{1}(a)}{\pi_e(a)} \big(r_a - V(\pi_1)\big)^2
- 0. 
\end{align*}
Compared with the form of IPW, one can find that the only difference is that the $r_a^2$ in the second term is replaced by $\big(r_a - V(\pi_1)\big)^2$, which is also related with the unknown parameters. 
Therefore, a tractable optimization objective can be formed by considering the upper bounds as 
\begin{align*}
    \argmin_{\pi_e}  \sum_{a \in [K]} \frac{\pi^2_{1}(a)}{\pi_e(a)}. 
\end{align*}
The other discussions on the safety constraint and the optimization for IPW can then be directly applied here. 
Similar results can be established for $V(\pi_1) - V(\pi_0)$, by noting the additivity. 

\subsection{Alternative objective function with side information for IPW}\label{sec:obj_side_IPW}
For MAB with IPW (similar arguments below apply for CMAB), recall that we derived the following form for the overall variance: 
\begin{align*}
T \times \var\Big(\widehat{V}_{IPW}(\pi_1; \mD_{e}) - \widehat{V}_{IPW}(\pi_0; \mD_{e}) \Big)
= 
\sum_{a \in [K]} \frac{\pi^2_{\Delta}(a)}{\pi_e(a)} \sigma^2_a 
+ \sum_{a \in [K]} \frac{\pi^2_{\Delta}(a)}{\pi_e(a)} r^2_a
- \big(V(\pi_1) - V(\pi_0) \big)^2, 
\end{align*}
Since either $\sigma_a$, $r_a$, or $V(\pi_1) - V(\pi_0)$ is known, this objective is intractable to directly optimize. 
From the worst-case point of view, we relax them using the upper bounds and optimize the  surrogate objective 
\begin{align*}
\sum_{a \in [K]} \frac{\pi^2_{\Delta}(a)}{\pi_e(a)}. 
\end{align*}
The advantage of this surrogate objective is that, for any instances, we can provide decent performance guarantee. 
However, the downside is such an objective might be conservative. 
In particular, even with side information (dataset $\mD_0$ or posterior distribution $Q$ over instance) available, our utilization of this information is limited (since the corresponding guarantee would be high-probability or Bayesian). 

We discuss alternative surrogate optimization objectives in this section. 
With the posterior distribution $Q$, 
it is easy to see that the objective, which is known as the Bayes risk \citep{brown1990information}, can be formulated as
\begin{align*}
&\Mean_Q \Big\{
\sum_{a \in [K]} \frac{\pi^2_{\Delta}(a)}{\pi_e(a)} \sigma^2_a 
+ \sum_{a \in [K]} \frac{\pi^2_{\Delta}(a)}{\pi_e(a)} r^2_a
- \big(V(\pi_1) - V(\pi_0) \big)^2
\Big\}\\
&= 
\sum_{a \in [K]} \frac{\pi^2_{\Delta}(a)}{\pi_e(a)} \Mean_Q\sigma^2_a 
+ \sum_{a \in [K]} \frac{\pi^2_{\Delta}(a)}{\pi_e(a)} \Mean_Qr^2_a
- \Mean_Q\big(V(\pi_1) - V(\pi_0) \big)^2
\\
&= 
\sum_{a \in [K]} \frac{\pi^2_{\Delta}(a)}{\pi_e(a)} (\Mean_Q\sigma^2_a +\Mean_Qr^2_a)
- \Mean_Q\big(V(\pi_1) - V(\pi_0) \big)^2, 
\end{align*}
which shares similar form with the one in the main text. 
Therefore, all discussions can be extended. 

Regarding the frequentist way, we can utilize $\mD_0$ to construct upper confidence bounds for the unknown terms, and the final performance guarantee would be a high-probability statement.

\section{Additional Experiment Details and Results}\label{sec:appendix_expt}

\subsection{Additional experiment details}\label{sec:more_expt_details}
In this section, we introduce more details of the experiment setup used in our simulation study. 
The configuration for CMAB is almost the same with that for MAB introduced in the main text, except for that we generate for every context independently. 
For linear bandits, we following \citet{kazerouni2016conservative} to make sure $\vx_i^T \vthe^*$ is positive for all arms when sampling the feature vectors. 
To generate the policies $\pi_0$ and $\pi_1$, we follow a similar design with MAB, except for that we first sample $\Tilde{\vthe}$  from the multivariate normal and then generate $\pi_1(a)$ (or $\pi_0(a)$) proportional to $\vx_a^T \Tilde{\vthe}$.




\subsection{Additional simulation experiments}\label{sec:more_expt}
\begin{figure}[!t]
  \centering
\begin{subfigure}{0.48\textwidth}
  \centering
  \includegraphics[width=\linewidth, height = 3.5cm]{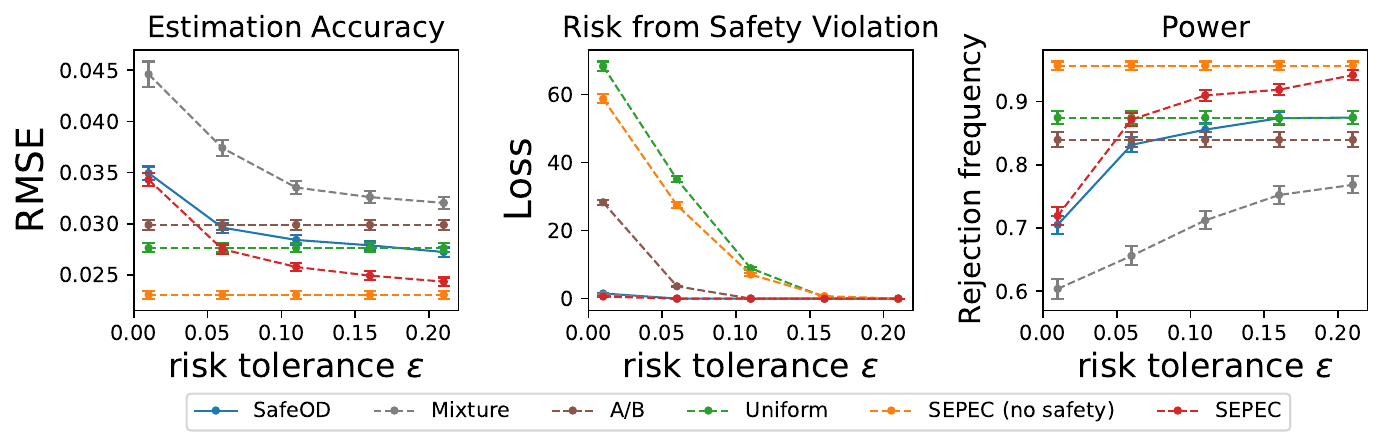}
  \caption{$T = 2000, K = 5, \sigma = 3$.}
\end{subfigure}%
\hspace{.5cm}
\begin{subfigure}{0.48\textwidth}
  \centering
  \includegraphics[width=\linewidth, height = 3.5cm]{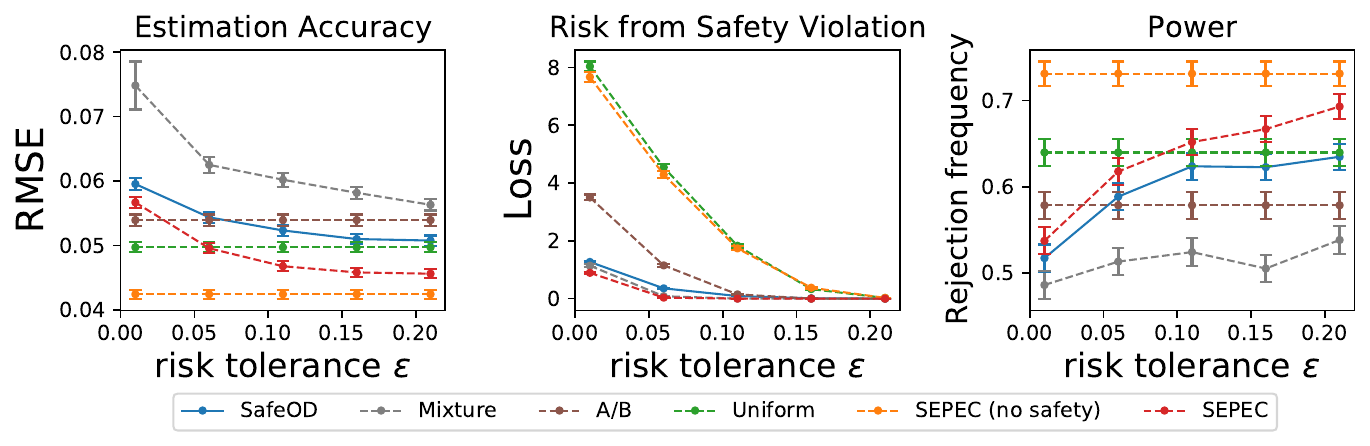}
  \caption{$T = 200, K = 20, \sigma = 2$.}
\end{subfigure}%
\caption{Experiment results for CMAB under other combinations of setting parameters.}\label{fig:more_CMAB}
\end{figure}

\begin{figure}[!t]
  \centering
\begin{subfigure}{0.48\textwidth}
  \centering
  \includegraphics[width=\linewidth, height = 3.5cm]{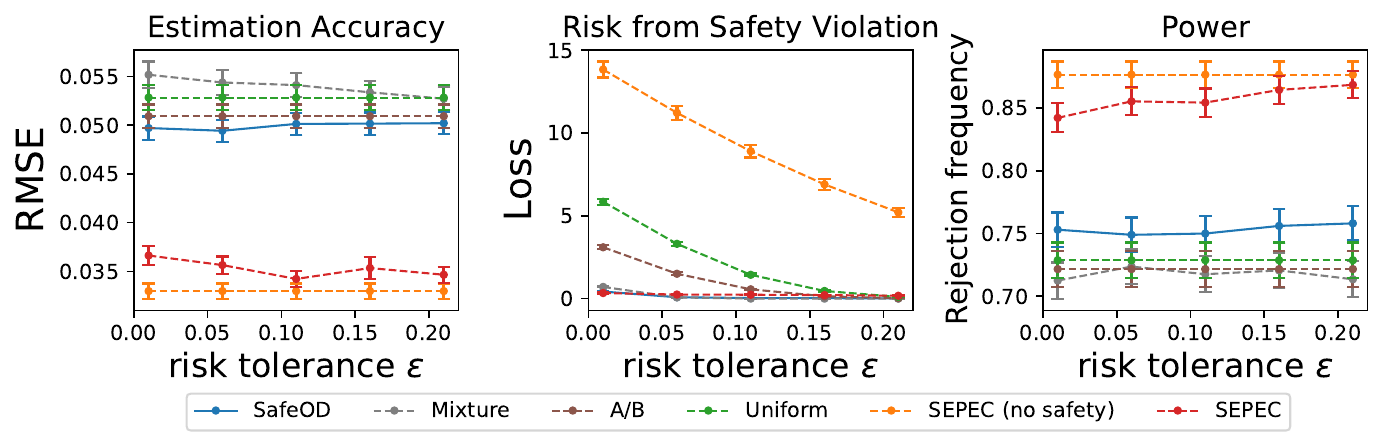}
  \caption{$T = 100, K = 100, \sigma = 2, d= 10$.}
\end{subfigure}%
\hspace{.5cm}
\begin{subfigure}{0.48\textwidth}
  \centering
  \includegraphics[width=\linewidth, height = 3.5cm]{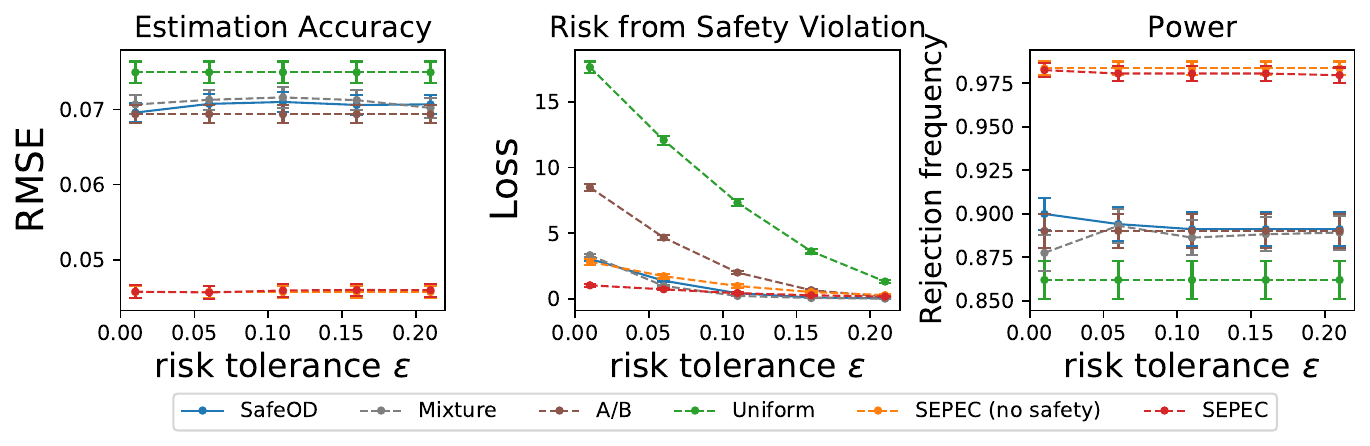}
  \caption{$T = 200, K = 100, \sigma = 2, d = 3$.}
\end{subfigure}%
\caption{Experiment results for LB under other combinations of setting parameters.}\label{fig:more_LB}
\end{figure}

In this section, we repeat the simulation experiments in the main text under a variety of parameter combinations to study the robustness of our findings. 
See Figure \ref{fig:more_CMAB} for results on CMAB and 
Figure \ref{fig:more_LB} for results on LB. 
Overall, the main findings are consistent with those in the main text.

\subsection{Experiments on the MNIST dataset}\label{sec:MNIST}
In this section, we conduct two experiments on the MNIST dataset \cite{deng2012mnist}. 
In this dataset, every arm is an  (vectorized) image of a digit between $0$ and $9$, 

\textbf{CMAB.}
The first experiment focuses on contextual bandits. 
As standard in the literature \cite{dudik2014doubly}, 
we adapt a classification task to a CMAB problem. 
Specifically, in every trail, we randomly pick $100$ images as our contexts, with their probability sampled from  $\text{Dirichlet}(\vone_{100})$. 
The $K = 10$ arms correspond to the $10$ digits, and only the arm corresponding to the one on the image returns reward $1$. 
All the other arms return reward $0$. 
The two policies $\pi_0$ and $\pi_1$ are two classifier trained with $1000$ randomly sampled data points, 
one using the decision tree and the other using multi-class logistic regression. 
We define the one with higher accuracy as $\pi_1$ when $H_1$ is true, and as $\pi_0$ when $H_0$ is true. 
We aim to collect $T = 1000$ data points. 

This setup aims to mimic a common application: we have two policies trained using logged data and with different functional form assumptions, so a direct comparison is unfair and may suffer from model misspecification. 
Therefore, one typically utilizes IPW (which is guaranteed to be unbiased under minimal assumptions) to compare these two policies. 


\textbf{Linear bandits.}
In the second experiment, we closely follow \citet{zhu2021safe} to study the performance on linear bandits. 
In every trail, we randomly pick one digit as the correct answer, and 
we assign reward $1$ to the images corresponding to this correct digit. 
$K = 100$ arms are randomly chosen. 
We set $\sigma = 1$, $|\mD_0| = 100$ and $T = 100$. 
$\vthe^*$ is fitted from the whole dataset. 

\textbf{Results.}
Results aggregated over $2000$ random seeds are presented in Figure \ref{fig:real}. 
The superior performance of \name{SEPEC} and other findings are largely consistent with our simulation experiments. 
In particularly, for linear bandits, the cost of satisfying the safety constraint is negligible, which is consistent with the findings in \citet{zhu2021safe}. 
Besides, regarding the power under linear bandits, we observe that  \name{SEPEC} is even slightly better than its non-safe version, though the difference is not statistically  significant. 
A closer look into the intermediate results tell that this is mainly due to that the FW is an iterative algorithm and the convergence situation might slightly vary due to numerical reasons. 


\section{Proofs}\label{sec:proofs}

\subsection{Proof of Lemma \ref{lemma:ipw_explicit_form}}
\begin{proof}
First, recall that 
\begin{align*}
T \times \nu(\pi_e; \vr, \{\sigma_a\})    
&=  T \times \var\Big(\widehat{V}_{IPW}(\pi_1; \mD_{e}) - \widehat{V}_{IPW}(\pi_0; \mD_{e}) \Big)\\
&= \sum_{a \in [K]} \frac{\pi^2_{\Delta}(a)}{\pi_e(a)} \sigma^2_a 
+ \sum_{a \in [K]} \frac{\pi^2_{\Delta}(a)}{\pi_e(a)} r^2_a
- \Big(V(\pi_1) - V(\pi_0)\Big)^2. 
\end{align*}

Let $W = \sum_a |\pi_{\Delta}(a)|$. 
We have $\pi_e^*(a) = |\pi_{\Delta}(a)| / W$. 
Therefore
\begin{equation*}
    \begin{split}
        &T \times \nu(\pi_e^*; \vr, \{\sigma_a\})\\
&= \sum_{a \in [K]} \frac{\pi^2_{\Delta}(a)}{|\pi_{\Delta}(a)| / W} \sigma^2_a 
+ \sum_{a \in [K]} \frac{\pi^2_{\Delta}(a)}{|\pi_{\Delta}(a)| / W} r^2_a
- c\\
&= \sum_{a \in [K]} |\pi_{\Delta}(a)|W (\sigma^2_a + r^2_a)
- \Big(V(\pi_1) - V(\pi_0)\Big)^2, 
    \end{split}
\end{equation*}
which implies
\begin{align*}
\nu(\pi_e^*; \vr, \{\sigma_a\})
=  
\frac{1}{T} \Big[
\sum_a |\pi_{\Delta}(a)| \times
\sum_{a \in [K]} |\pi_{\Delta}(a)| (\sigma^2_a + r^2_a)
- \big(V(\pi_1) - V(\pi_0)\big)^2
\Big]
\end{align*}

\end{proof}

\subsection{Proof of Theorem \ref{thm:mab}}
\begin{proof}
First, recall that 
\begin{align*}
T \times \nu(\pi_e; \vr, \{\sigma_a\})    
= T \times \var\Big(\widehat{V}_{IPW}(\pi_1; \mD_{e}) - \widehat{V}_{IPW}(\pi_0; \mD_{e}) \Big)
= \sum_{a \in [K]} \frac{\pi^2_{\Delta}(a)}{\pi_e(a)} \sigma^2_a 
+ \sum_{a \in [K]} \frac{\pi^2_{\Delta}(a)}{\pi_e(a)} r^2_a
- c, 
\end{align*}
where $c = \Big(V(\pi_1) - V(\pi_0) \Big)^2$ is independent with $\pi_e$.

Denote $\Mean_Q(r^2_a) \equiv r^2_Q$ and $\Mean_Q(\sigma^2(a)) \equiv \sigma^2_Q$. 
Therefore, 
for any $\pi_e \in \Pi$, 
we have 
\begin{align*}
&T \times \Mean_{Q} \Big[ \nu(\pi_e^*; \vr, \{\sigma_a\}) - \nu(\pi_e; \vr, \{\sigma_a\})   \Big]\\
&= \Mean_{Q} \Big[ \sum_{a \in [K]} \frac{\pi^2_{\Delta}(a)}{\pi_e^*(a)} \sigma^2_a 
+ \sum_{a \in [K]} \frac{\pi^2_{\Delta}(a)}{\pi_e^*(a)} r^2_a
- \sum_{a \in [K]} \frac{\pi^2_{\Delta}(a)}{\pi_e(a)} \sigma^2_a 
- \sum_{a \in [K]} \frac{\pi^2_{\Delta}(a)}{\pi_e(a)} r^2_a
\Big]\\
&= 
\Big[ \sum_{a \in [K]} \frac{\pi^2_{\Delta}(a)}{\pi_e^*(a)}
- \sum_{a \in [K]} \frac{\pi^2_{\Delta}(a)}{\pi_e(a)}
\Big] \times \big(\sigma^2_a + r^2_a\big)\\
&\le 0, 
\end{align*}
where the last inequality is due to that, by design, $\pi_e^*$ is the minimizer of $\sum_{a \in [K]} \frac{\pi^2_{\Delta}(a)}{\pi_e(a)}$ within the class of safe policies. 


To prove the minimax optimality, we first show that, for any policy $\pi_e$, 
$\max_{(\vr, {\sigma_a})  \in \mathcal{I}} \nu(\pi_e; \vr, {\sigma_a})$ is achieved when $\vr = \vone$ and $\sigma_a \equiv \sigma$. 
To see this, we exam
\begin{align*}
&T \times \nu(\pi_e; \vr, \{\sigma_a\}) - T \times \nu(\pi_e; \vone, \{\sigma, \dots, \sigma\}) \\
&=\Big[
\sum_{a \in [K]} \frac{\pi^2_{\Delta}(a)}{\pi_e(a)} \sigma^2_a 
+ \sum_{a \in [K]} \frac{\pi^2_{\Delta}(a)}{\pi_e(a)} r^2_a
- (V(\pi_1) - V(\pi_0))^2
\Big]\\
&\quad-
\Big[
\sum_{a \in [K]} \frac{\pi^2_{\Delta}(a)}{\pi_e(a)} \sigma^2
+ \sum_{a \in [K]} \frac{\pi^2_{\Delta}(a)}{\pi_e(a)} 1^2
- 0^2
\Big]\\
&=\sum_{a \in [K]} \frac{\pi^2_{\Delta}(a)}{\pi_e(a)} (\sigma^2_a - \sigma^2)
+ \sum_{a \in [K]} \frac{\pi^2_{\Delta}(a)}{\pi_e(a)} (r^2_a-1)
- (V(\pi_1) - V(\pi_0))^2\\
&\le 0.  
\end{align*}

Therefore, we have 
\begin{align*}
T \times \max_{(\vr, {\sigma_a})  \in \mathcal{I}} \nu(\pi_e; \vr, {\sigma_a})
= \sum_{a \in [K]} \frac{\pi^2_{\Delta}(a)}{\pi_e(a)} (\sigma^2 + 1)
\end{align*}
Notice that $\pi_e^*$, by design, is the minimizer of this objective. 
Therefore, we have that $\pi_e^*$ is minimax optimal, i.e., 
$ \max_{(\vr, {\sigma_a}) \in \mathcal{I}}  \nu(\pi_e^*; \vr, {\sigma_a}) = \min_{\pi_e \in \Pi}  \max_{(\vr, {\sigma_a})  \in \mathcal{I}} \nu(\pi_e; \vr, {\sigma_a})$. 





\end{proof}

\subsection{Proof of Theorem \ref{thm:LB}}\label{sec:proof_LB}
\begin{proof}
Denote $\vG(\pi_e) = \sum_{\vx \in \mathcal{A}} \pi_e(\vx) \vx \vx^T. $
We first note that 
\begin{align*}
T \times    \nu\pi_e;\mD_0) 
    = T \times \bar{\vphi}^T_{\Delta \pi}
(T\vG(\pi_e)+ \vPhi_0^{T}\vPhi_0)^{-1}
\bar{\vphi}_{\Delta \pi}
    = \bar{\vphi}^T_{\Delta \pi}
(\vG(\pi_e)+ T^{-1}\vPhi_0^{T}\vPhi_0)^{-1}
\bar{\vphi}_{\Delta \pi}
\to \bar{\vphi}^T_{\Delta \pi}\vG(\pi_e)^{-1}\bar{\vphi}_{\Delta \pi}, 
\end{align*}
when $T$  goes to infinity. 
Besides, we note the following relationship
\begin{align*}
    T \times \nu^*(\pi_e; \mD_0) 
    &= T \times \var_{\vPhi \sim \pi_e} \Big(\widehat{V}_{DM}(\pi_1; \mD_{\vPhi} \cup \mD_0) - \widehat{V}_{DM}(\pi_0; \mD_{\vPhi} \cup \mD_0) \Big)\\
    &=  \Mean_{\vPhi \sim \pi_e}\Big[\bar{\vphi}^T_{\Delta \pi}
     (T^{-1}\vPhi^T\vPhi+ T^{-1}\vPhi_0^{T}\vPhi_0)^{-1}
    \bar{\vphi}_{\Delta \pi}
 \Big].
\end{align*}

Since the rows of $\vPhi$ are \textit{i.i.d.}, 
 by the strong law of large numbers, 
we know $T^{-1}\vPhi^T\vPhi \overset{a.s.}{\to} \vG(\pi_e)$, 
and hence $T^{-1}\vPhi^T\vPhi + T^{-1}\vPhi_0^{T}\vPhi_0 \overset{a.s.}{\to} \vG(\pi_e)$. 
Therefore, by the continuous mapping theorem, we have $(T^{-1}\vPhi^T\vPhi + T^{-1}\vPhi_0^{T}\vPhi_0)^{-1} \overset{a.s.}{\to} \vG(\pi_e)^{-1}$ and also 
$\bar{\vphi}^T_{\Delta \pi} \big[T^{-1}\vPhi^T\vPhi + T^{-1}\vPhi_0^{T}\vPhi_0\big]^{-1} \bar{\vphi}_{\Delta \pi}
\overset{a.s.}{\to} \bar{\vphi}^T_{\Delta \pi}\vG(\pi_e)^{-1}\bar{\vphi}_{\Delta \pi}$. 
Finally, from the random matrix theory for the tail bounds on the eignevalues of random matrix with $i.i.d.$ rows  \cite{vershynin2010introduction}, we know the uniformly integrable condition can be satisfied and hence we conclude with 
\begin{align*}
    \Mean_{\vPhi \sim \pi_e} \Big[
    \bar{\vphi}^T_{\Delta \pi} \big[T^{-1}\vPhi^T\vPhi + T^{-1}\vPhi_0^{T}\vPhi_0\big]^{-1} \bar{\vphi}_{\Delta \pi}
    \Big] \to \bar{\vphi}^T_{\Delta \pi}\vG(\pi_e)^{-1}\bar{\vphi}_{\Delta \pi}
\end{align*}





\end{proof}

\subsection{Counter-example}\label{sec:Counter-example}
In this section, we argue that there does not exist a policy $\pi_e$ that dominates all other policies across all problem instances, when we use the IPW estimator. 

Suppose $\pi_1 \neq \pi_0$. 
For any policy $\pi_e$, we can always construct an instance $(\vr, \{\sigma_a\})$ and a policy $b'$, such that \begin{align*}
    \nu(\pi_e; \vr, \{\sigma_a\}) >  \nu(\pi_e'; \vr, \{\sigma_a\}). 
\end{align*}

To see this, note that 
\begin{align*}
    T \times \Big[ \nu(\pi_e; \vr, \{\sigma_a\}) -  \nu(\pi_e'; \vr, \{\sigma_a\}) \Big]
= 
\sum_{a \in [K]} \frac{\pi^2_{\Delta}(a)}{\pi_e(a)} \sigma^2_a 
+ \sum_{a \in [K]} \frac{\pi^2_{\Delta}(a)}{\pi_e(a)} r^2_a
- 
\sum_{a \in [K]} \frac{\pi^2_{\Delta}(a)}{b'(a)} \sigma^2_a 
- \sum_{a \in [K]} \frac{\pi^2_{\Delta}(a)}{b'(a)} r^2_a. 
\end{align*}
By setting $r_a \equiv 0$, we get 
\begin{align*}
    T \times \Big[ \nu(\pi_e; \vr, \{\sigma_a\}) -  \nu(\pi_e'; \vr, \{\sigma_a\}) \Big]
= 
\sum_{a \in [K]} 
\big(
\frac{1}{\pi_e(a)}
- \frac{1}{b'(a)}
\big)
\pi^2_{\Delta}(a) \sigma^2_a. 
\end{align*}
Since $\pi_1 \neq \pi_0$, there are at least two arms $a'$ and $a''$ where the two policies differ. 
By design, we know $\pi_e(a')$ and $\pi_e(a'')$ are both positive. 
A counter-example can hence be designed by setting $\sigma_a' = 0$, $\sigma_{a''} = 1$, $b'(a') = 0.5\pi_e(a')$,  $b'(a'') = 0.5\pi_e(a') + \pi_e(a'')$ and keeping $\pi_e(a) = b'(a)$ on the other arms. 
In other words, we move more budgets to the arm of high variance. 
More precisely, we have 
\begin{align*}
&T \times \Big[ \nu(\pi_e; \vr, \{\sigma_a\}) -  \nu(\pi_e'; \vr, \{\sigma_a\}) \Big]\\
&= 
0 + 
\big(
\frac{1}{\pi_e(a')}
- \frac{1}{b'(a')}
\big)
\pi^2_{\Delta}(a') \sigma^2_{a'}
+ 
\big(
\frac{1}{\pi_e(a'')}
- \frac{1}{b'(a'')}
\big)
\pi^2_{\Delta}(a'') \sigma^2_{a''}\\
&= 
\big(
\frac{1}{\pi_e(a'')}
- \frac{1}{0.5\pi_e(a') + \pi_e(a'')}
\big)
\pi^2_{\Delta}(a'') \sigma^2_{a''}\\
&> 0. 
\end{align*}

\subsection{Convexity of the objectives and constraints}\label{sec:convexity}
For the sake of completeness, we show in this section that the objectives and constraints considered in this paper are all convex. 
Regarding the objective of IPW, we note the relationship that 
\begin{align*}
     \alpha\sum_{a \in [K]} \frac{\pi^2_{\Delta}(a)}{\pi_e(a)}
    + (1-\alpha) \sum_{a \in [K]} \frac{\pi^2_{\Delta}(a)}{\pi'_e(a)}
    \ge  \sum_{a \in [K]} \frac{\pi^2_{\Delta}(a)}{\alpha \pi_e(a) + (1-\alpha)\pi'_e(a)}, 
\end{align*}
due to the inequality
\begin{align*}
     \alpha \frac{1}{\pi_e(a)}
    + (1-\alpha)  \frac{1}{\pi'_e(a)}
    \ge   \frac{1}{\alpha \pi_e(a) + (1-\alpha)\pi'_e(a)}, \forall a \in [K], 
\end{align*}
which stems from the convexity of $f(x) = 1/x$. 
The same arguments hold for CMAB with IPW. 

Regarding the objective of linear bandits with DM, 
\begin{align*}
\bar{\vphi}^T_{\Delta \pi}
    (T\sum_{\vx \in \mathcal{A}} \pi_e(\vx) \vx \vx^T+ \vPhi_0^{T}\vPhi_0)^{-1}
    \bar{\vphi}_{\Delta \pi}, 
\end{align*}
we notice that $\sum_{\vx \in \mathcal{A}} \pi_e(\vx) \vx \vx^T$  is linear in $\pi_e$ and the matrix inverse operator is a convex function. 
Therefore, their composition is still convex.

Regarding the constraint, notice that, for every single instance, the constraint is a linear one and hence convex. 
The overall feasible set is the intersection of these convex sets, and hence is convex.

\end{document}